\documentclass[3p,11pt]{elsarticle}

\journal{Knowledge-Based Systems}

\usepackage{amssymb}
\usepackage{amsmath}
\usepackage{bbm}
\usepackage{stmaryrd}
\usepackage{environ}
\usepackage{algorithmic}
\usepackage{algorithm}
\usepackage{url}
\usepackage{graphicx}
\usepackage[caption=false,font=footnotesize]{subfig}
\usepackage{multirow}

\usepackage{color}
\usepackage{ulem}
\newcommand{\reels}{\mathbb{R}}

\newcommand{\Proba}{\mathbb{P}}

\newcommand{\calA}{{\cal A}}

\newcommand{\calN}{{\cal N}}

\newcommand{\betah}{\widehat{\beta}}

\def\balpha{\boldsymbol{\alpha}}

\def\bc{\boldsymbol{c}}

\def\negtheta#1{\overline{\stheta#1}}
\def\stheta#1{\{\theta_#1\}}

\newcommand{\bi}{\begin{itemize}}
\newcommand{\ei}{\end{itemize}}
\newcommand{\be}{\begin{enumerate}}
\newcommand{\ee}{\end{enumerate}}

\NewEnviron{eqs}[1]
{
\begin{subequations}
\label{#1}
\begin{align}
\BODY
\end{align}
\end{subequations}
}

\newtheorem{Prop}{Proposition}  
\newtheorem{Rem}{Remark}

\begin{document}

\begin{frontmatter}

\title{Logistic Regression, Neural Networks and Dempster-Shafer Theory: \\
a New Perspective}
 
\author{Thierry œDen{\oe}ux\corref{cor1}}
\ead{thierry.denoeux@utc.fr}

\address{Universit\'e de Technologie de Compi\`egne, CNRS \\
UMR 7253 Heudiasyc, Compi\`egne, France}

\begin{abstract}
We revisit  logistic regression and its nonlinear extensions, including multilayer feedforward neural networks, by showing that these classifiers can be viewed as converting input or higher-level features into Dempster-Shafer mass functions and aggregating them by Dempster's rule of combination. The probabilistic outputs of these classifiers are the normalized plausibilities corresponding to the underlying combined mass function. This mass function is more informative than the output probability distribution. In particular, it makes it possible to distinguish between lack of evidence (when none of the features provides discriminant information) from conflicting evidence (when different features support different classes). This expressivity of mass functions allows us to gain insight into the role played by each input feature in logistic regression, and to interpret hidden unit outputs in multilayer neural networks. It also makes it possible to use alternative decision rules, such as  interval dominance, which select a set of classes when the available evidence does not unambiguously point to a single class, thus trading reduced error rate for higher imprecision.
\end{abstract}


\end{frontmatter}

\section{Introduction}
\label{sec:intro}

The Dempster-Shafer (DS) theory of belief functions \cite{dempster67a} \cite{shafer76} is now well-established as a formalism for reasoning and making decisions with uncertainty \cite{yager08}. DS theory, also referred to as \emph{Evidence Theory}, is essentially based on representing independent pieces of evidence by completely monotone capacities (also called belief functions), and pooling them using a generic operator called Dempster's rule of combination. 

In the last twenty years, DS theory has been increasingly applied to statistical pattern recognition and, in particular, to supervised classification. One direction of research is \emph{classifier fusion}: classifier outputs are expressed as belief functions and combined by Dempster's rule or any other rule (see, e.g., \cite{xu92,rogova94,quost11,bi12,liu18,jiang17}). Another approach is \emph{evidential calibration}, which converts  the decisions of statistical classifiers (such as support vector machines)  into belief functions \cite{xu16a,minary17,liu19}. The third approach, which is maybe the most promising and the focus of this paper,  is to design \emph{evidential classifiers}, whose basic principles are rooted in DS theory. Typically, an evidential classifier breaks down the evidence of each input feature vector  into elementary mass functions and   combines them by Dempster's rule. The combined mass function  (or \emph{orthogonal sum}) can then be used for decision-making \cite{denoeux96b}. Thanks to the generality and expressiveness of the belief function formalism, evidential classifiers provide more informative outputs than those of conventional classifiers. This expressiveness can be exploited, in particular, for uncertainty quantification, novelty detection and information fusion in decision-aid or fully automatic decision systems.

Over the years, several principles for designing  evidential classifiers have been developed. In \cite{denoeux06c}, a distinction was made between the so-called \emph{model-based} approach, which uses estimated class-conditional distributions and the ``Generalized Bayes Theorem'', an extension of Bayes theorem \cite{smets93b,appriou91}, and the \emph{case-based}, or \emph{distance-based} approach, in which mass functions $m_j$ are constructed based on distances to learning instances or to prototypes. Evidential  classifiers in the latter category have been used in a wide range of applications \cite{su09,guettari16,chen18}. They include the \emph{evidential $k$-nearest neighbor  rule} \cite{denoeux95a} and its variants (see, e.g. \cite{jiao13,liu13a,lian15,lian16,su18}),  as well as the \emph{evidential neural network classifier} \cite{denoeux00a}, in which mass functions are constructed based on the distances to  prototypes, and the whole system is trained to minimize an error function.


In this paper, we show that not only these particular model-based and distance-based classifiers, but also a  broad class of supervised machine learning algorithms, can be seen as evidential classifiers. This class contains logistic regression and its non linear generalizations, including multilayer feedforward neural networks, generalized additive models, support vector machines and, more generally, all classifiers based on linear combinations of input or higher-order features and their transformation through the logistic or softmax transfer function. We will show that \emph{generalized logistic regression classifiers} can be seen as computing the orthogonal sum of elementary pieces of evidence supporting each class or its complement. The output class probabilities are then normalized plausibilities according to some underlying Dempster-Shafer mass function, the expression of which is laid bare in this paper.  This ``hidden'' mass function provides a more informative description of the classifier output than the class probabilities, and can be used for decision-making. Also, the individual mass functions computed by each of the features provides insight into the internal operation of classifier and can help to interpret its decisions. This finding leads us to the conclusion that DS theory is a much more general framework for classifier analysis and construction than was initially believed, and opens a new perspective for the study and practical application of a wide range of machine learning algorithms\footnote{A preliminary version of this paper with some partial results appeared as a short conference paper \cite{denoeux18a}.}.

The rest of this paper is organized as follows. DS theory and some principles of classifier construction will first be recalled in Section \ref{sec:background}. The new connection between DS theory and some machine learning models will then be established  in Section \ref{sec:main}, and the identification of DS model will be addressed in Section \ref{sec:ident}.  Finally, some numerical experiments  will be presented in Section \ref{sec:results}, and Section \ref{sec:concl} will conclude the paper.

\section{Background}
\label{sec:background}

In this section, we first recall some necessary definitions and results from DS theory (Section \ref{subsec:DS}). We then provide brief descriptions of logistic regression and neural network classifiers that will be considered later in the paper (Section \ref{subsec:classif}).

\subsection{Dempster-Shafer theory}
\label{subsec:DS}

\subsubsection{Mass function}

Let $\Theta=\{\theta_1,\ldots,\theta_K\}$ be a finite set. A \emph{mass function} on $\Theta$ is a mapping $m: 2^\Theta \rightarrow [0,1]$ such that $m(\emptyset)=0$ and 
\[
\sum_{A\subseteq \Theta} m(A)=1.
\]
In DS theory, $\Theta$ is the set of possible answers to some question, and a mass function $m$ represent a piece of evidence pertaining to that question. Each mass $m(A)$ represents   a  share  of  a  unit  mass  of  belief  allocated  to the hypothesis that the truth is in $A$, and  which  cannot  be  allocated  to  any  strict  subset  of $A$. Each subset $A\subseteq \Theta$ such that $m(A)>0$ is called a \emph{focal set} of $m$. A mass function $m$ is said to be \emph{simple} if it has the following form:
\begin{equation}
\label{eq:simple}
m(A)=s, \quad m(\Theta)=1-s,
\end{equation}
for some $A\subset \Theta$ such that $A\neq \emptyset$ and some $s\in [0,1]$, called the \emph{degree of support} in $A$. For a reason that will become apparent later, the quantity $w:=-\ln(1-s)$ is called the \emph{weight of  evidence}\footnote{This notion of ``weight of evidence'' in DS theory should not be confused with related, but different notions with similar names proposed in other contexts such as rough set theory, as reviewed in \cite{ko16}.} associated to $m$ \cite[page 77]{shafer76}. The \emph{vacuous} mass function, corresponding to $s=w=0$, represents complete ignorance.

\subsubsection{Belief and Plausibility functions}

Given a mass function $m$, \emph{belief} and \emph{plausibility} functions are defined, respectively, as follows:
\begin{subequations}
\begin{align}
\label{eq:belief}
Bel(A)&:=\sum_{ B \subseteq A} m(B)\\
Pl(A)&:=\sum_{B \cap A \neq \emptyset} m(B)=1-Bel(\overline{A}),
\end{align}
\end{subequations}
for all $A\subseteq \Theta$. The quantity $Bel(A)$ can be interpreted as the degree of total support to $A$, while $1-Pl(A)$ is the degree of total support to  $\overline{A}$, i.e., the degree of doubt in $A$ \cite{shafer76}. The \emph{contour function} $pl:\Theta\rightarrow [0,1]$ is the restriction of the plausibility function $Pl$ to singletons, i.e., $pl(\theta)=Pl(\{\theta\})$, for all $\theta\in\Theta$. 

\subsubsection{Dempster's rule}

Two mass functions $m_1$  and $m_2$ representing independent items of evidence can be combined using  Dempster's rule \cite{dempster67a,shafer76} defined as 
\begin{equation}
\label{eq:dempster}
(m_1 \oplus m_2)(A):=\frac{1}{1-\kappa} \sum_{B \cap C=A} m_1(B) m_2(C),
\end{equation}
for all $A\subseteq \Theta$, $A\neq\emptyset$, and $(m_1 \oplus m_2)(\emptyset):=0$. In (\ref{eq:dempster}),  $\kappa$ is the \emph{degree of conflict} between the two mass functions, defined as 
\begin{equation}
\label{eq:conflict}
\kappa:=\sum_{B \cap C=\emptyset} m_1(B) m_2(C).
\end{equation}
Mass function $m_1 \oplus m_2$ is well defined if $\kappa<1$. It is then called the \emph{orthogonal sum} of $m_1$ and $m_2$. Dempster's rule is commutative and associative, and the vacuous mass function is its only neutral element. The contour function $pl_1\oplus pl_2$ associated to $m_1\oplus m_2$  can be computed as
\begin{equation}
\label{eq:contourDS}
pl_1\oplus pl_2(\theta) = \frac{pl_1(\theta) pl_2(\theta)}{1-\kappa}, 
\end{equation}
for all $\theta  \in \Theta$.

\subsubsection{Weights of evidence}

Given two simple mass functions $m_1$ and $m_2$ with the same focal set $A$ and degrees of support $s_1$ and $s_2$, their orthogonal sum is the simple mass function
\begin{eqs}{dempster_simple}
(m_1 \oplus m_2)(A)&=1-(1-s_1)(1-s_2)\\
(m_1 \oplus m_2)(\Theta)&=(1-s_1)(1-s_2).
\end{eqs}
The corresponding weight of evidence is, thus, 
\begin{eqs}{wdempster}
w&=-\ln[(1-s_1)(1-s_2)]\\
&=-\ln(1-s_1)-\ln(1-s_2)=w_1+w_2,
\end{eqs}
i.e., weights of evidence add up when aggregating evidence using Dempster's rule. Denoting a simple mass function with focal set $A$ and weight of evidence $w$ as $A^w$, this property can be expressed by the following equation,
\begin{equation}
\label{eq:dempsterw}
A^{w_1}\oplus A^{w_2}=A^{w_1+w_2}.
\end{equation}
We note that, in \cite{denoeux08}, following \cite{smets95c}, we used the term ``weight'' for $-\ln w$. As we will see, the additivity property is central in our analysis: we thus stick to Shafer's terminology and notation in this paper. A mass function is said to be \emph{separable} if it can be decomposed as the orthogonal sum of simple mass functions \cite[page 87]{shafer76}. A separable mass function can thus be written as
\[
m= \bigoplus_{\emptyset\neq A \subset \Theta} A^{w(A)},
\]
where $w(\cdot)$ is a mapping from $2^\Theta\setminus \{\emptyset,\Theta\}$ to $[0,+\infty)$.

\subsubsection{Plausibility Transformation}

It is sometimes useful to approximate  a DS mass function $m$ by a probability mass function $p_m:\Theta \rightarrow [0,1]$. One such approximation with good properties is obtained by normalizing the contour function \cite{voorbraak89,cobb06}; we then have
\begin{equation}
\label{eq:plaustrans}
p_m(\theta_k):= \frac{pl(\theta_k)}{\sum_{l=1}^K pl(\theta_l)}, \quad k=1,\ldots,K.
\end{equation}
As a consequence of (\ref{eq:contourDS}), the so-called \emph{plausibility transformation} (\ref{eq:plaustrans}) has the following interesting property in relation with Dempster's rule:
\[
p_{m_1\oplus m_2}(\theta_k) \propto p_{m_1}(\theta_k)  p_{m_2}(\theta_k), \quad k=1,\ldots,K,
\]
i.e., the probability distribution associated to $m_1\oplus m_2$ can be computed in $O(K)$ arithmetic operations  by multiplying the probability distributions $p_{m_1}$ and $p_{m_2}$ elementwise, and renormalizing.

\subsubsection{Least Commitment Principle}
\label{subsubsec:LCP}

The  \emph{maximum uncertainty} \cite{klir99} or \emph{least commitment} \cite{smets93b} principle serves the same purpose as the maximum entropy principle in probability theory. According to this principle, when several belief functions are compatible with a set of constraints, the least committed (or informative) should be selected. In order to apply this principle, we need to define a partial order on the set of belief functions. For that purpose, we may either define a degree of imprecision or  uncertainty of a belief function \cite{klir99}, or we may adopt a more qualitative approach and directly define an {informational} ordering relation on the set of belief functions \cite{dubois86a,yager86b}. 

If we restrict ourselves to separable mass functions, as will be done in this paper, we can compare mass functions by their weights of evidence. Given two separable mass functions $m_1= \bigoplus_{\emptyset\neq A \subset \Theta} A^{w_1(A)}$ and $m_2= \bigoplus_{\emptyset\neq A \subset \Theta} A^{w_2(A)}$, it makes sense to consider that $m_1$ is more committed than $m_2$ (denoted as $m_1\sqsubseteq_w m_2$) if it has larger weights of evidence, i.e, if $w_1(A) \ge w_2(A)$ for all  $A$ \cite{denoeux08}. Because of (\ref{eq:dempsterw}), combining $m_2$ with a separable mass function $m$ results in a more committed mass function $m_1=m_2\oplus m$, with $m_1 \sqsubseteq_w m_2$. 

A related family of measures of information content is defined by
\begin{equation}
\label{eq:Ip}
I_p(m):=\sum_{\emptyset\neq A \subset \Theta} w(A)^p, \quad p>0.
\end{equation}
Clearly, for any two separable mass functions $m_1$ and $m_2$, $m_1\sqsubseteq_w m_2 \Rightarrow I_p(m_1)\ge I_p(m_2)$.

 \subsubsection{Decision Analysis}
\label{subsubsec:decision}
Consider a decision problem with a set $\calA=\{a_1,\ldots,a_r\}$ of acts, a set $\Theta=\{\theta_1,\ldots,\theta_K\}$ of states of nature, and a loss function $L:\calA\times\Theta \rightarrow \reels$. The lower and upper risks of act $a$ with respect to a mass function $m$ are defined, respectively, as the lower and upper expected loss \cite{dempster67a,shafer81}, if the decision-maker (DM)  selects act $a$:
\begin{align*}
R_*(a):=\sum_{A\subseteq \Theta} m(A) \min_{\theta\in A} L(a,\theta),\\
R^*(a):=\sum_{A\subseteq \Theta} m(A) \max_{\theta\in A} L(a,\theta).
\end{align*}
A pessimistic (resp., optimistic) DM will prefer act $a$ over $a'$ if $R^*(a)\le R^*(a')$ (resp.,  $R_*(a)\le R_*(a')$). Alternatively, a conservative approach is to consider $a$ preferable to $a'$ whenever $R^*(a)\le R_*(a')$. This \emph{interval dominance (ID)} preference relation  \cite{troffaes07} is a partial preorder on $\calA$. For decision-making, one can select the set of maximal elements of this relation, defined as $\{a\in \calA \vert \forall a'\in \calA \setminus\{a\}, R^*(a') > R_*(a)\}$. In classification, act $a_k$ is usually interpreted as selecting class $k$, and we have $r=K$. Assuming the 0-1 loss function defined by $L(a_k,\theta_l)=1-\delta_{kl}$, where $\delta$ is the Kronecker delta, we have $R_*(a_k)=1-pl(\theta_k)$ and $R^*(a_k)=1-Bel(\{\theta_k\})$. The optimistic rule then selects the class with the highest plausibility \cite{denoeux96b}. This rule will be hereafter referred to as the \emph{maximum plausibility (MP)} rule.

\subsection{Logistic Regression}
\label{subsec:classif}

In the following, we recall some basic definitions and notations about classification. We start with binary logistic regression and proceed with the multi-category case and  some nonlinear extensions.

\subsubsection{Binary Logistic Regression}

Consider a binary classification problem with $d$-dimensional feature vector $X=(X_1,\ldots,X_d)$ and class variable $Y\in \Theta=\{\theta_1,\theta_2\}$. Let $p_1(x)$ denote the probability that $Y=\theta_1$ given that $X=x$. In the binary logistic regression model, it is assumed that
\begin{equation}
\label{eq:logreg}
\ln \frac{p_1(x)}{1-p_1(x)}=\beta^T x + \beta_0,
\end{equation}
where $\beta \in \reels^d$ and $\beta_0\in \reels$ are parameters. Solving (\ref{eq:logreg}) for $p_1(x)$, we get
\begin{equation}
\label{eq:logistic}
p_1(x)=\frac{1}{1+\exp[-(\beta^T x+ \beta_0)]}.
\end{equation}
Given a learning set $\{(x_i,y_i)\}_{i=1}^n$, parameters $\beta$ and $\beta_0$ are usually estimated by maximizing the conditional log-likelihood
\begin{equation}
\label{eq:loglik}
\ell(\beta,\beta_0)=\sum_{i=1}^n  y_{i1}\ln p_1(x_i) + (1-y_{i1})\ln \left[1-p_1(x_i)\right],
\end{equation}
where $y_{i1}=1$ if $y_i=\theta_1$ and $y_{i1}=0$ otherwise.

\subsubsection{Multinomial logistic regression}
\label{subsubsec:nonlinear}

Consider now a multiclass classification problem with $K>2$ classes, and let $\Theta=\{\theta_1,\ldots,\theta_K\}$ denote the set of classes. Multinomial logistic regression extends binary logistic regression by assuming the log-posterior probabilities to be affine functions of $x$:
\begin{equation}
\ln p_k(x)=\beta_k^T x+\beta_{k0}+\gamma, \quad k=1,\ldots,K,
\end{equation}
 where $p_k(x)=\Proba(Y=\theta_k \vert X=x)$ is the posterior probability of class $\theta_k$, $\beta_k\in \reels^d$ and $\beta_{k0} \in \reels$ are class-specific parameters and $\gamma \in \reels$ is a constant that does not depend on $k$. The posterior probability of class $\theta_k$ can then be expressed as
\begin{equation}
\label{eq:softmax}
p_k(x)=\frac{\exp(\beta_k^T x+\beta_{k0})}{\sum_{l=1}^K \exp(\beta_l^T x+\beta_{l0})},
\end{equation}
and parameters $(\beta_k,\beta_{k0})$, $k=1\ldots,K$ can be estimated by maximizing the conditional likelihood as in the binomial case. The transformation from linear combinations of features $\beta_k^T x+\beta_{k0} \in \reels$ to probabilities in $[0,1]$ described by (\ref{eq:softmax}) is often referred to as the \emph{softmax transformation}.

\subsubsection{Nonlinear extensions} 

\begin{figure}
\centering  
\includegraphics[width=0.7\textwidth]{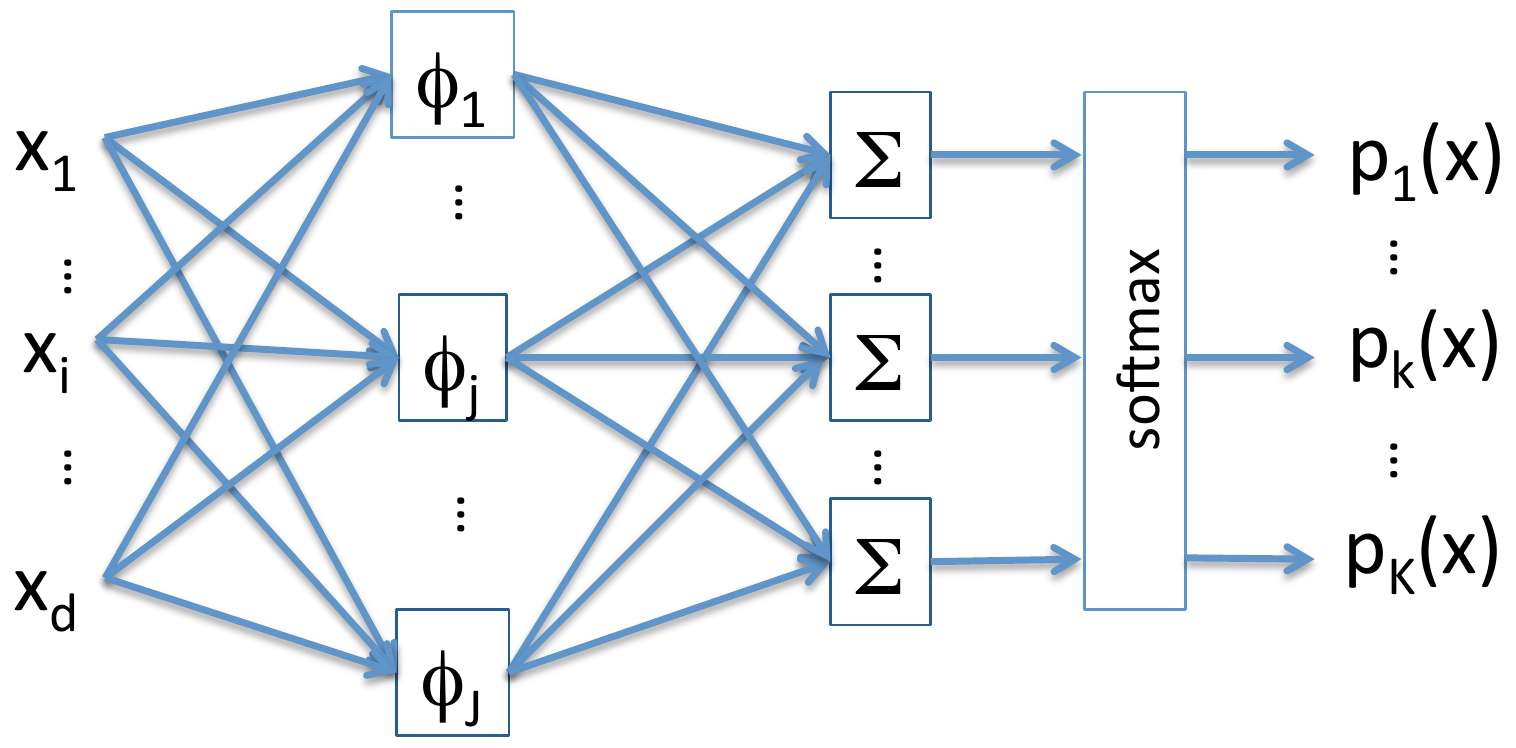}
\caption{Generalized logistic regression classifier. \label{fig:generalized_logistic}}
\end{figure}

Logistic regression classifiers define decision regions separated by hyperplanes: they are linear classifiers. However, nonlinear classifiers can be built by applying logistic regression to transformed features $\phi_j(x)$, $j=1,\ldots,J$, where the $\phi_j$'s are nonlinear mappings from $\reels^d$ to $\reels$. We call such classifiers \emph{generalized logistic regression (GLR) classifiers} (see Figure \ref{fig:generalized_logistic}). Both the new features $\phi_j(x)$ and the coefficients $(\beta_k,\beta_{k0})$ are usually learnt simultaneously by minimizing some cost function. Popular models based on this principle include quadratic logistic regression \cite{jiang19}, generalized additive models \cite{hastie90}, multilayer feedforward neural networks \cite{rumelhart86,goodfellow16}, radial basis function networks \cite{moody89} and support vector machines \cite{sholkopf02}. 
In particular, Feedforward Neural Networks (FNNs) are models composed of elementary computing units (or ``neurons'') arranged in layers. Each layer computes a vector of new features as functions of the outputs from the previous layer. 
For classification, the output layer is typically a softmax layer with $K$ output units. This model is thus equivalent to logistic regression performed on new features computed in the network's hidden layers. 
All weights in the network are learnt by minimizing a cost function, which is often taken as the negative conditional likelihood (or cross-entropy), as in logistic regression.

\section{DS analysis of GLR classifiers}
\label{sec:main}

In this section, we expose the main result of this paper, which establishes a bridge between DS theory, recalled in Section \ref{subsec:DS}, and the GLR classifiers summarized in Section \ref{subsec:classif}. We start with  binary classification  in Section \ref{subsec:K2}, and proceed with the multi-category case in Section \ref{subsec:Kg2}.

\subsection{Case $K=2$}
\label{subsec:K2}

Consider a binary classification problem with $K=2$ classes in $\Theta=\{\theta_1,\theta_2\}$. Let $\phi(x)=(\phi_1(x),\ldots,\phi_J(x))$ be a vector of $J$ features. These features may be the input features, in which case we have $\phi_j(x)=x_j$ for all $j$ and $J=d$, or nonlinear functions thereof. Each feature value $\phi_j(x)$ is a piece of evidence about the class $Y \in \Theta$ of the instance under consideration. Assume that this evidence points either to $\theta_1$ or $\theta_2$, depending on the sign of
\begin{equation}
\label{eq:weights2}
w_j:=\beta_j \phi_j(x) + \alpha_j,
\end{equation}
where $\beta_j$ and $\alpha_j$ are two coefficients. The weights of evidence for $\theta_1$ and $\theta_2$ are assumed to be equal to, respectively, the positive part $w_j^+:=\max(0,w_j)$ of $w_j$, and  its negative part $w_j^-:=\max(0,-w_j)$. Under this model, the consideration of feature $\phi_j$ induces the simple mass function 
\[
m_j=\stheta{1}^{w_j^+} \oplus \stheta{2}^{w_j^-}.
\]

\subsubsection{Output mass function}

\begin{figure}
\centering  
\subfloat[$m(\{\theta_1\})$ \label{fig:mtheta1}]{\includegraphics[width=0.4\textwidth]{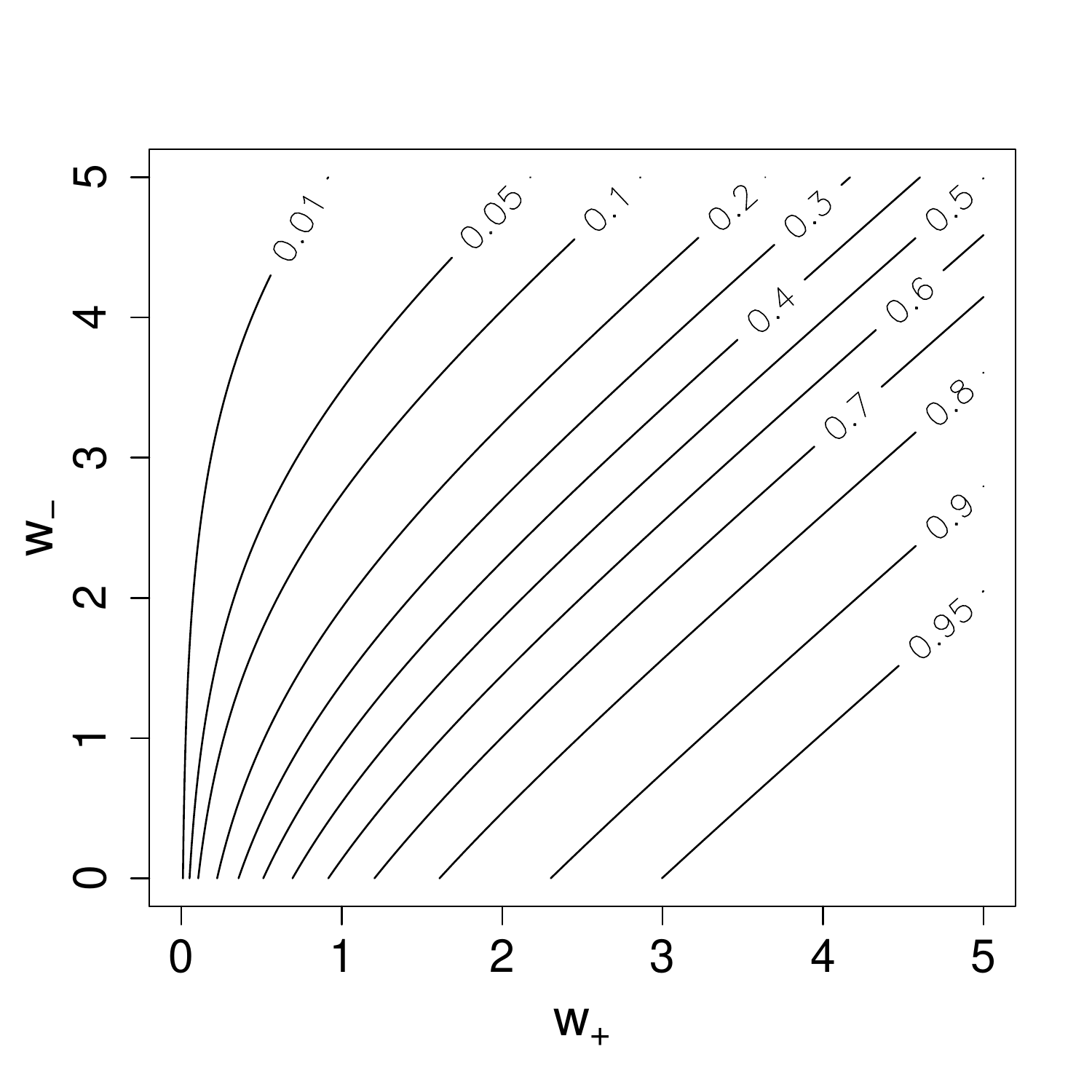}}
\subfloat[$m(\Theta)$\label{fig:mTheta}]{\includegraphics[width=0.4\textwidth]{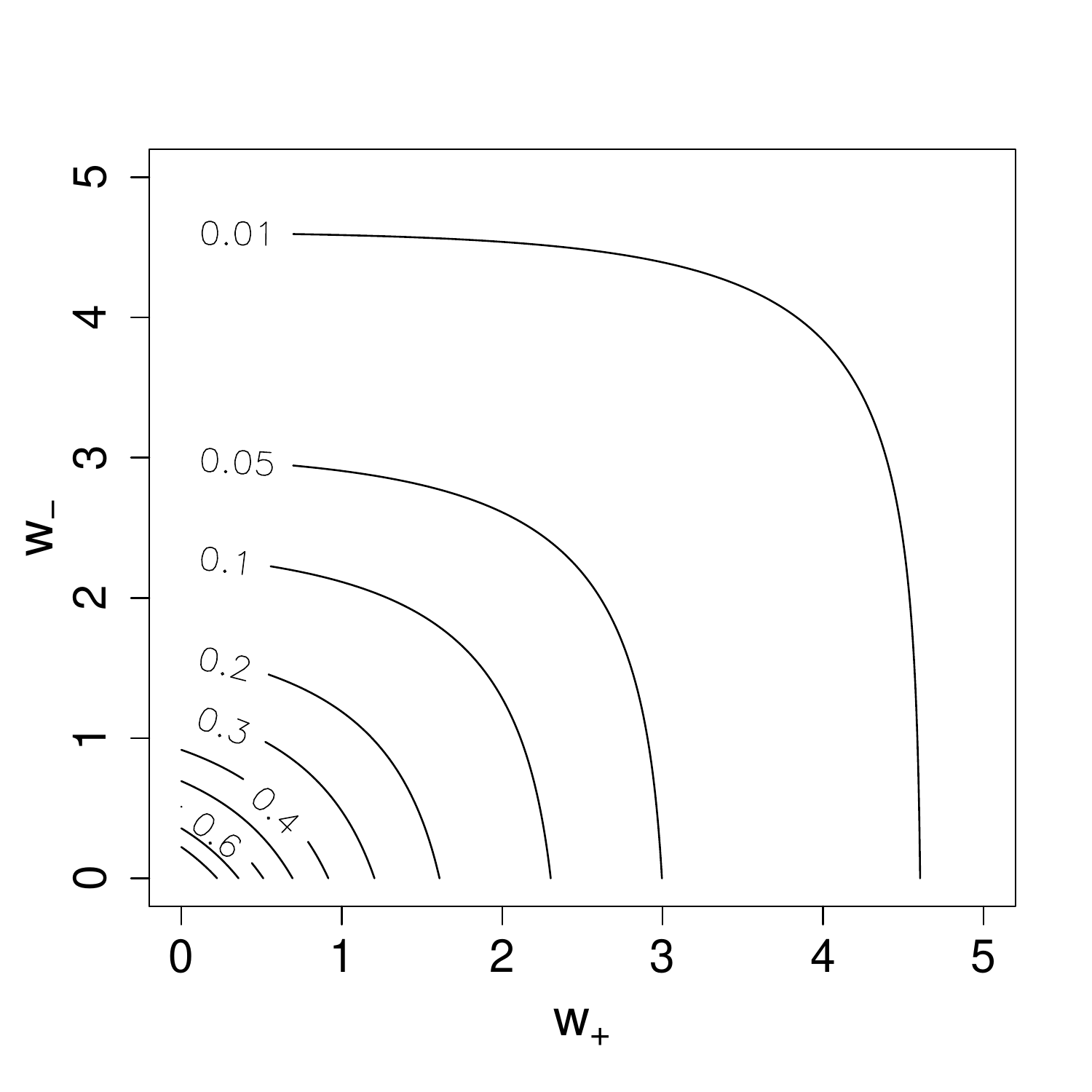}}\\
\subfloat[Degree of conflict $\kappa$ \label{fig:conflict}]{\includegraphics[width=0.4\textwidth]{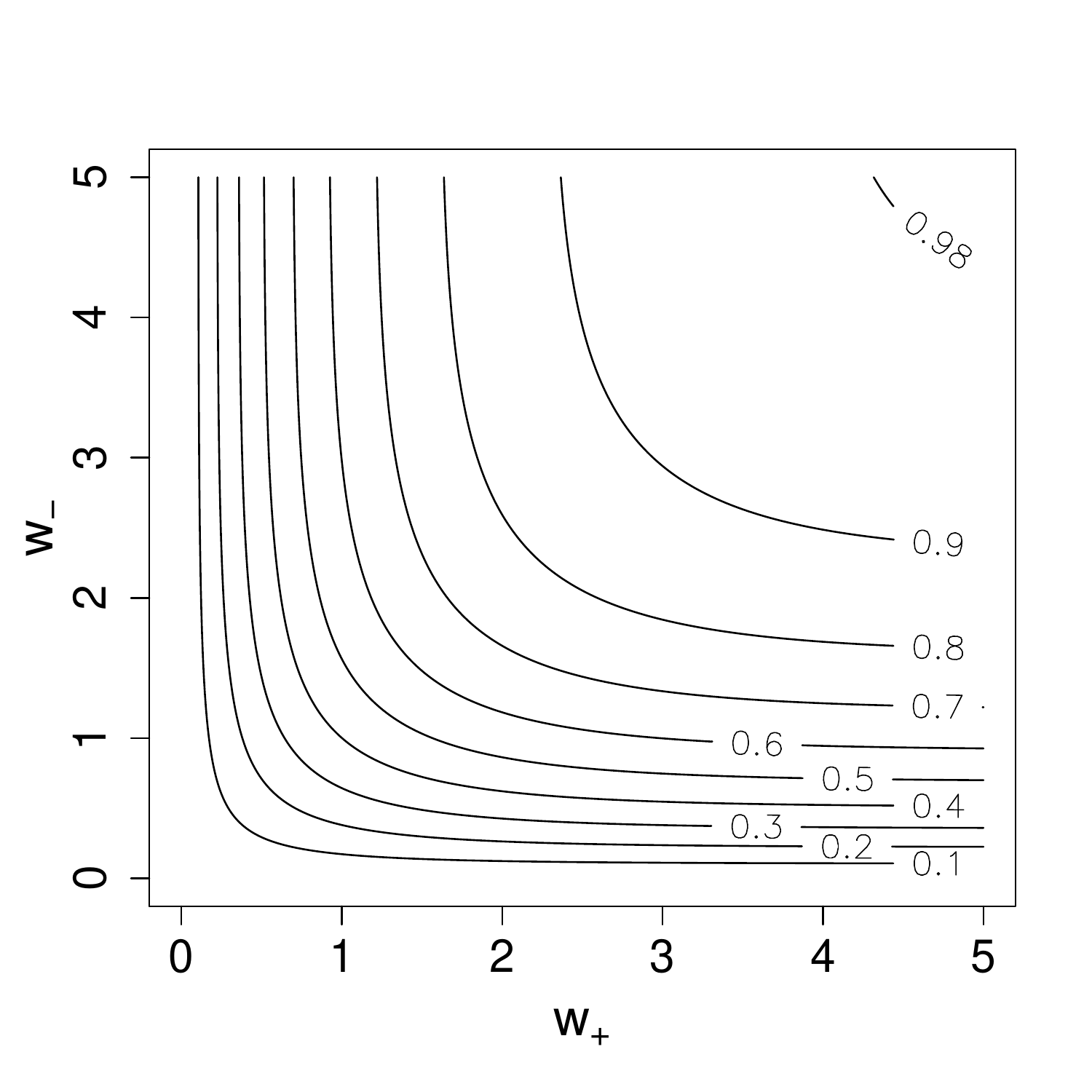}}
\caption{Contour lines of $m(\{\theta_1\})$ (a), $m(\Theta)$ (b) and the degree of conflict $\kappa$ (c) as functions of $w^+$ (horizontal axis) and $w^-$ (vertical axis). \label{fig:masses}}
\end{figure}

Assuming  that the values of the $J$ features can be considered as independent pieces of evidence, the combined mass function after taking into account the $J$ features is 
\begin{eqs}{m12def}
m &= \bigoplus_{j=1}^J \left( \stheta{1}^{w_j^+} \oplus \stheta{2}^{w_j^-}\right)\\
&= \left(\bigoplus_{j=1}^J  \stheta{1}^{w_j^+}\right) \oplus\left(\bigoplus_{j=1}^J \stheta{2}^{w_j^-}\right)\\
\label{eq:mpm}
&=\stheta{1}^{w^+} \oplus \stheta{2}^{w^-},
\end{eqs}
where $w^+:=\sum_{j=1}^J w_j^+$  and $w^-:=\sum_{j=1}^J w_j^-$  are the total weights of evidence supporting, respectively, $\theta_1$ and $\theta_2$. Denoting by $m^+$ and $m^-$ the two mass functions on the right-hand side of Eq. (\ref{eq:mpm}), we have
\begin{eqs}{m12p}
m^+(\stheta{1}) &= 1-\exp(-w^+)\\
m^+(\Theta) &= \exp(-w^+)
\end{eqs}
and
\begin{eqs}{m12m}
m^-(\stheta{2}) &= 1-\exp(-w^-)\\
m^-(\Theta) &= \exp(-w^-).
\end{eqs}
Hence,
\begin{eqs}{eq:m12}
m(\stheta{1}) &= \frac{[1-\exp(-w^+)]\exp(-w^-)}{1-\kappa} \\
m(\stheta{2}) &= \frac{[1-\exp(-w^-)]\exp(-w^+)}{1-\kappa}\\
m(\Theta) &= \frac{\exp(-w^+-w^-)}{1-\kappa}=\frac{\exp(-\sum_{j=1}^J|w_j|)}{1-\kappa},
\end{eqs}
where
\begin{equation}
\label{eq:conf12}
\kappa=[1-\exp(-w^+)] [1-\exp(-w^-)]
\end{equation}
is the degree of conflict between $m^-$ and $m^+$. Mass function $m$ defined Eqs (\ref{eq:m12}) and (\ref{eq:conf12}) is the output of the evidential classifier. As shown in Figure \ref{fig:mtheta1}, $m(\stheta{1})$ is  increasing w.r.t. $w^+$ and decreasing w.r.t. $w^-$, while the mass $m(\Theta)$ is a decreasing function of the total weight of evidence $w^++w^-$ (Figure \ref{fig:mTheta}). The degree of conflict $\kappa$ increases with both $w^-$ and $w^+$ (Figure \ref{fig:conflict}).

\subsubsection{Contour function}

The contour function corresponding to $m$ is
\begin{eqs}{eq:pltheta2}
pl(\theta_1) &= m(\stheta{1})+m(\Theta)=\frac{\exp(-w^-)}{1-\kappa} \\
pl(\theta_2) &= m(\stheta{2})+m(\Theta)=\frac{\exp(-w^+)}{1-\kappa}.
\end{eqs}
We can observe that Eq. (\ref{eq:pltheta2}) is consistent with the semantics of plausibility: the plausibility of class $\theta_1$ is high when there is little evidence in favor of $\theta_2$, i.e., when $w^-$ is low.  Applying the plausibility transformation (\ref{eq:plaustrans}), we get the following probability of $\theta_1$:
\begin{eqs}{eq:ptheta1}
p_m(\theta_1)&=\frac{\exp(-w^-)}{\exp(-w^-)+\exp(-w^+)}\\
&=\frac{1}{1+\exp(w^- -w^+)}\\
\label{eq:ptheta1_3}
&= \frac{1}{1+\exp[-(\beta^T\phi(x)+\beta_0)]},
\end{eqs}
with $\beta=(\beta_1,\ldots,\beta_J)$ and 
\begin{equation}
\label{eq:constr_alpha2}
\beta_0=\sum_{i=1}^J \alpha_j.
\end{equation}

\subsubsection{Discussion}

\begin{figure}
\centering  
\includegraphics[width=0.6\textwidth]{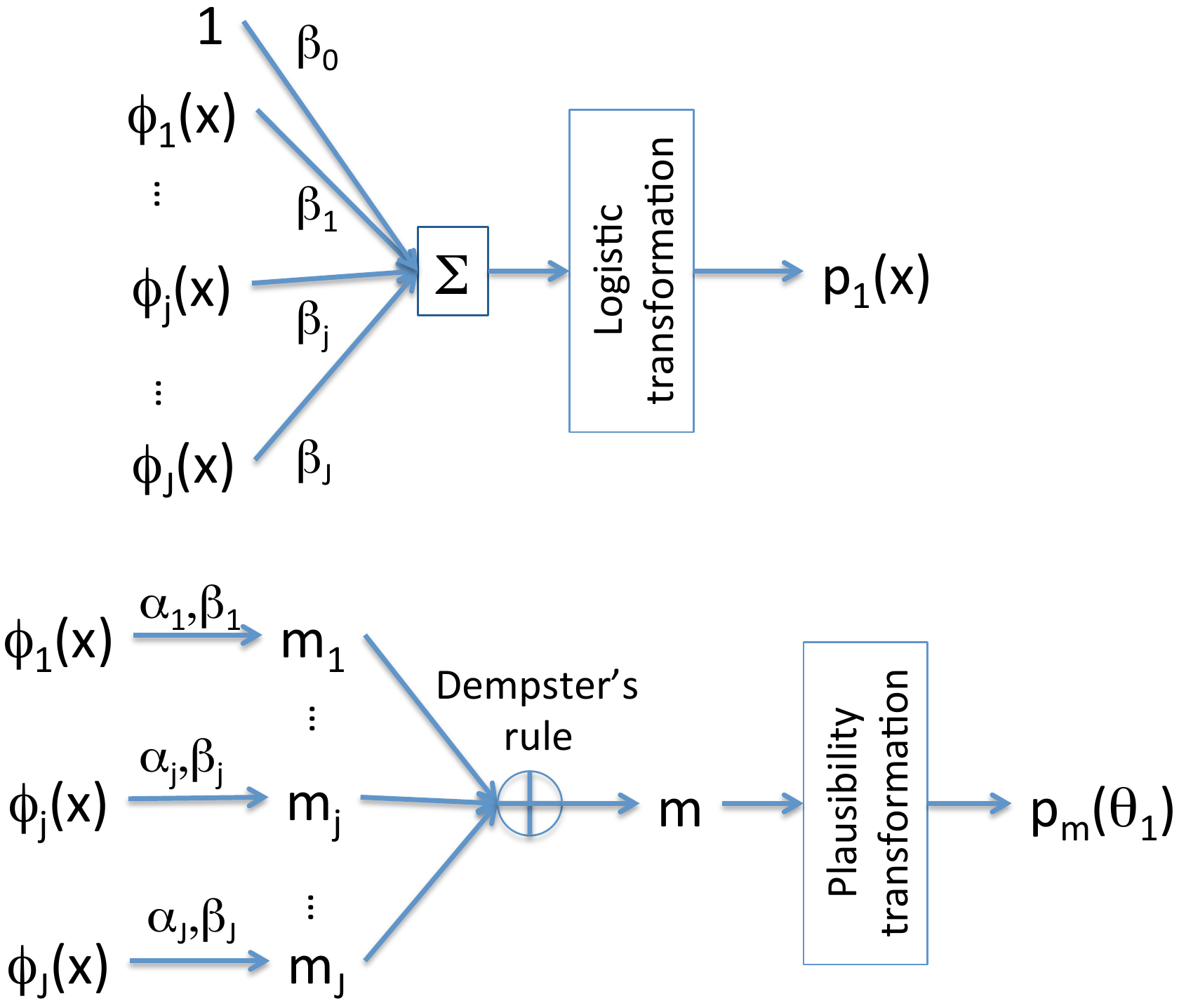}
\caption{Classical view (top) and DS view (bottom) of a binary GLR classifier. \label{fig:evidential_logistic_binary}}
\end{figure}

We observe that (\ref{eq:ptheta1_3}) is identical to (\ref{eq:logistic}): in the two-category case, the probabilities computed by logistic regression can, thus, be viewed as normalized plausibilities obtained by a process of evidence combination in the DS framework.  Figure \ref{fig:evidential_logistic_binary} contrasts the classical view of binary logistic regression with the DS view outlined here. If one considers only the output probability, both views are equivalent. The latter, however, lays bare an underlying mass function $m$, which  has one more degree of freedom than  the output probability $p_1(x)$. This additional degree of freedom makes it possible to distinguish, e.g.,  between lack of evidence, in which case we have $m(\Theta)=1$, and maximally conflicting evidence  corresponding to $m(\stheta{1})= m(\stheta{2})=0.5$. These two cases result in the same output probability $p_1(x)=0.5$. This distinction has implications for decision making, as will be shown in Section \ref{sec:results}.

Typically, parameters $\beta$ and $\beta_0$ are estimated by maximizing the log-likelihood function (\ref{eq:loglik}), but parameters $\alpha_j$, $j=1,\ldots,J$ are unidentifiable. We will return to this point in Section \ref{sec:ident}. Before that, we address the multi-category case in the following section.

\subsection{Case $K>2$}
\label{subsec:Kg2}

Let us now consider the multi-category case, where $K>2$. 

\subsubsection{Model}

For each $\theta_k$, we now assume that the evidence of feature $\phi_j(x)$ points either to the singleton $\stheta{k}$ or to its complement $\negtheta{k}$, depending on the sign of
\begin{equation}
\label{eq:weights1}
w_{jk}:=\beta_{jk} \phi_j(x) + \alpha_{jk},
\end{equation}
where $(\beta_{jk},\alpha_{jk})$, $k=1,\ldots,K$, $j=1,\ldots,J$ are parameters. The weights of evidence for $\stheta{k}$ and $\negtheta{k}$ are supposed to be equal, respectively, to the positive and negative parts of $w_{jk}$, denoted by $w^+_{jk}$ and $w^-_{jk}$, respectively.
 For each feature $\phi_j$ and each class $\theta_k$, we thus have two simple mass functions, $m_{jk}^+:=\stheta{k}^{w_{jk}^+}$ and $m_{jk}^-:=\negtheta{k}^{w_{jk}^-}$.
Assuming these mass functions to be independent, they can be combined by Dempster's rule. Combining separately the positive and the negative evidence with respect to each class $\theta_k$, we get
\begin{eqs}{mkpm}
\label{eq:mkp}
m_k^+&:=\bigoplus_{j=1}^J m^+_{jk} =\stheta{k}^{w_{k}^+} \\
\label{eq:mkm}
m_k^-&:=\bigoplus_{j=1}^J m^-_{jk} =\negtheta{k}^{w_{k}^-},
\end{eqs}
where
\begin{equation}
\label{eq:weights3}
w_k^+:=\sum_{j=1}^J w^+_{jk} \quad \text{and} \quad w_k^-:=\sum_{j=1}^J w^-_{jk}.
\end{equation}

\subsubsection{Combined contour function}
\begin{figure}
\centering  
\subfloat[]{ \includegraphics[width=0.5\textwidth]{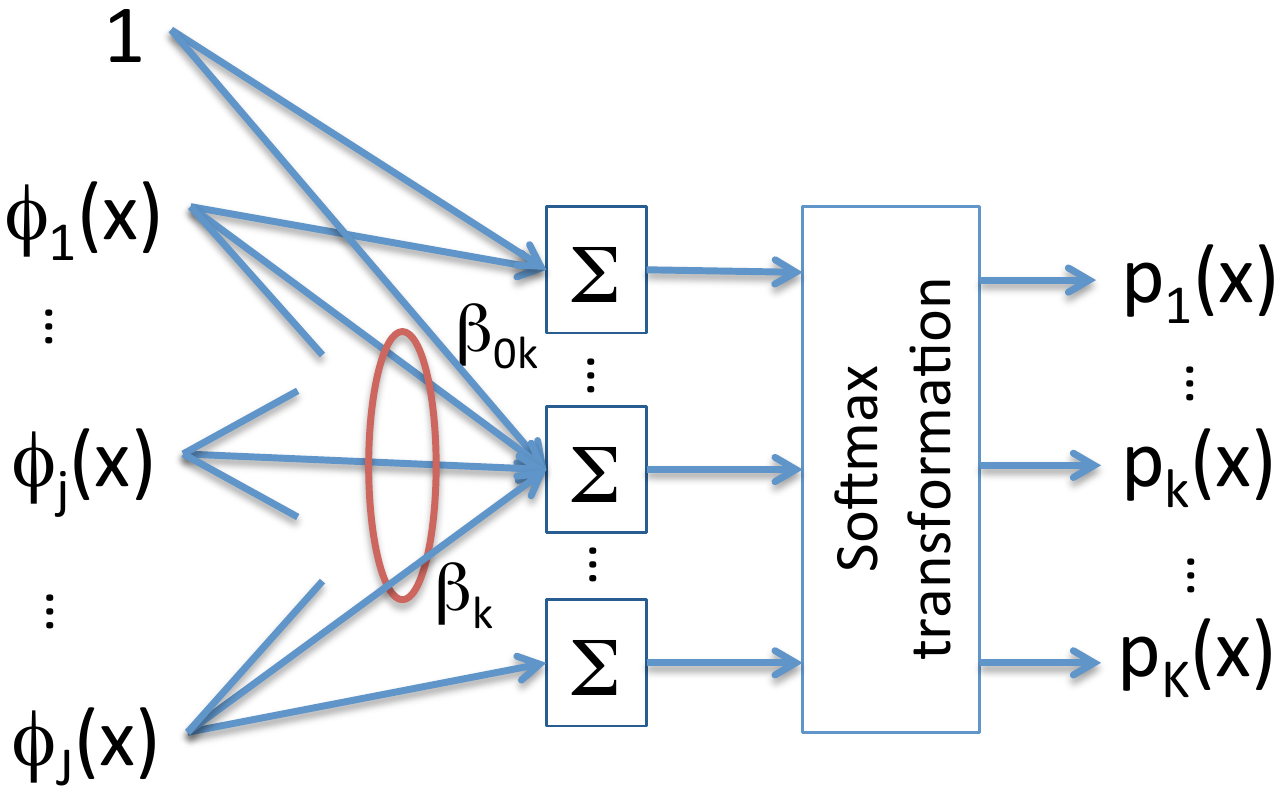}}\\
\subfloat[]{\includegraphics[width=0.6\textwidth]{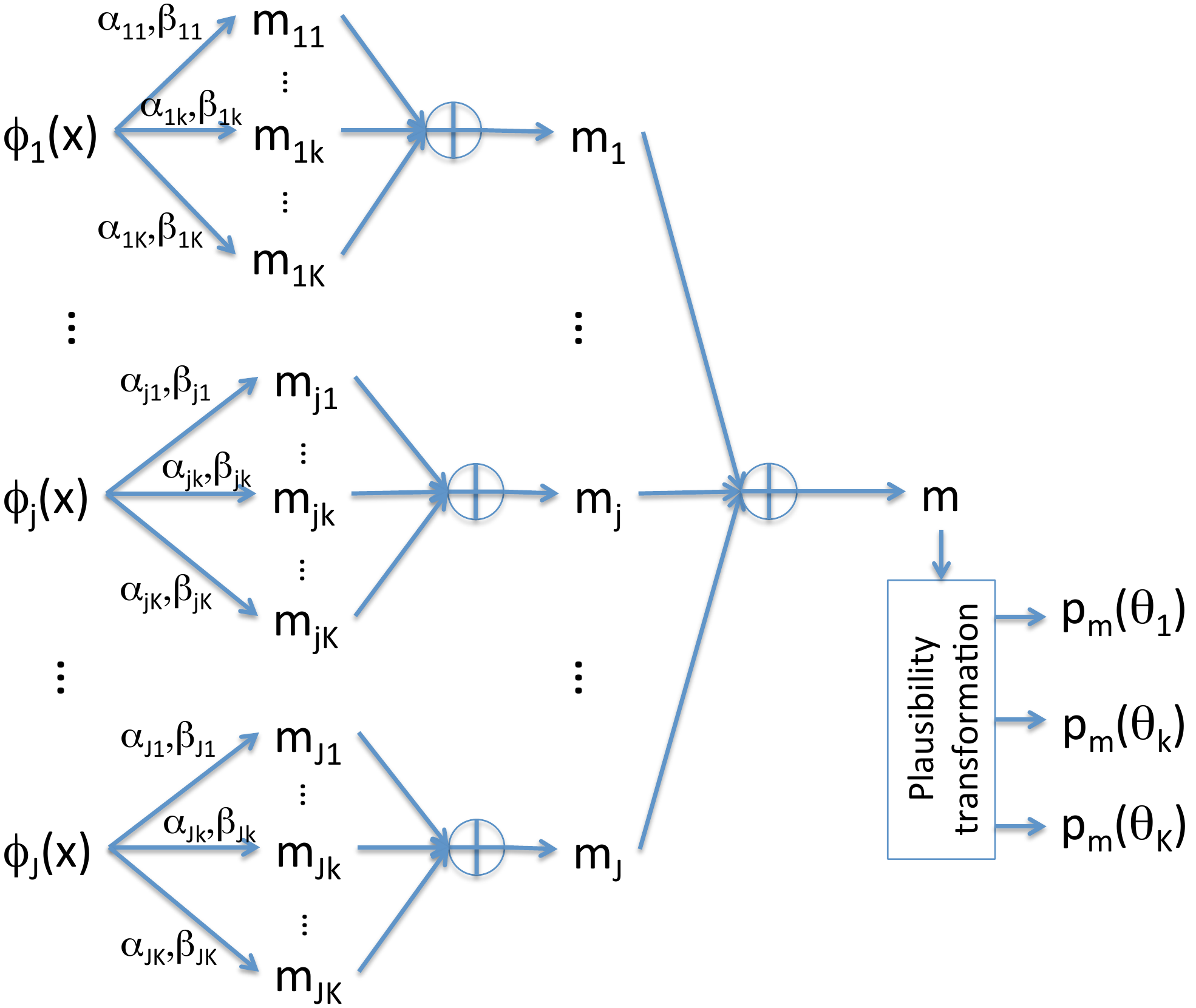}}
\caption{Classical view (a) and DS view (b) of a multinomial GLR classifier. \label{fig:evidential_logistic_K}}
\end{figure}

The contour functions $pl_k^+$ and  $pl_k^-$ associated, respectively, with $m_k^+$ and $m_k^-$ are
\[
pl_k^+(\theta)=\begin{cases}
1 & \text{if } \theta=\theta_k,\\
\exp\left(-w_{k}^+\right) & \text{otherwise,}
\end{cases}
\]
and 
\[
pl_k^-(\theta)=\begin{cases}
\exp\left(-w_{k}^-\right) & \text{if } \theta=\theta_k,\\
1 & \text{otherwise.}
\end{cases}
\] 
Now, let $m^+=\bigoplus_{k=1}^K m^+_{k}$ and $m^-=\bigoplus_{k=1}^K m^-_{k}$ be the mass functions pooling, respectively, all the positive and the negative evidence, and let $pl^+$ and $pl^-$ be the corresponding contour functions. From (\ref{eq:contourDS}), we have
\begin{equation*}
pl^+(\theta_k)  \propto \prod_{l=1}^K pl^+_l(\theta_k)= 
\exp\left(-\sum_{l\neq k} w_l^+\right)=\exp\left(-\sum_{l=1}^K w_l^+\right) \exp(w_k^+)
\propto \exp(w_k^+),
\end{equation*}
and
\begin{equation}
\label{eq:plm}
pl^-(\theta_k)  \propto \prod_{l=1}^K pl^-_l(\theta_k)= \exp(-w_k^-).
\end{equation}
Finally, let $m=m^+\oplus m^-$ and let $pl$ be the corresponding contour function. Using again Eq. (\ref{eq:contourDS}), we have
\begin{equation*}
pl(\theta_k) \propto pl^+(\theta_k)pl^-(\theta_k)\propto \exp(w_k^+ -w_k^-) \propto 
\exp\left(\sum_{j=1}^J w_{jk}\right)= 
\exp\left(\sum_{j=1}^J \beta_{jk}\phi_j(x) + \sum_{j=1}^J \alpha_{jk}\right).
\end{equation*}
Let $p$ be the probability mass function induced from $m$ by the plausibility transformation (\ref{eq:plaustrans}), and let
\begin{equation}
\label{eq:beta_alpha}
\beta_{0k}:=\sum_{j=1}^J \alpha_{jk}. 
\end{equation}
The probability of class $\theta_k$ induced by mass function $m$ is
\begin{equation}
\label{eq:normpl}
p_m(\theta_k)=\frac{\exp\left(\sum_{j=1}^J \beta_{jk}\phi_j(x)+\beta_{0k}\right)}{\sum_{l=1}^K \exp\left(\sum_{j=1}^J\beta_{jl}\phi_j(x)+\beta_{0l}\right)}.
\end{equation}
It is identical to (\ref{eq:softmax}). We thus have proved that the result found in Section \ref{subsec:ident2} for the binary case also holds in the multi-category case: conditional class probabilities computed by a multinomial GLR classifier can be seen as the normalized plausibilities obtained after combining elementary mass functions $m_{jk}=m_{jk}^+\oplus m_{jk}^-$ by Dempster's rule: these classifiers are, thus, evidential classifiers as defined in Section \ref{sec:intro}. The classical and DS views of multinomial GLR classifiers are contrasted in Figure \ref{fig:evidential_logistic_K}.

\subsubsection{Output mass function}

As in the binary case, we can compute the expression of the underlying mass function $m$. Its expression in the multi-category case is more complex than it is in the binary case. It is given in the following proposition.

\begin{Prop}
\label{prop:mass}
The output mass function
\begin{equation}
\label{eq:m}
m= \bigoplus_{k=1}^K \left(\stheta{k}^{w_{k}^+} \oplus \negtheta{k}^{w_{k}^-}\right)
\end{equation}
has the following expression:
\begin{equation*}
m(\stheta{k})= C \exp(-w_k^-) \left\{ \exp(w_k^+)-1 + \prod_{l\neq k} \left[1-\exp(-w_l^-)\right]\right\},
\end{equation*}
for $k=1,\ldots,K$, and 
\begin{equation*}
m(A) = 
C \left\{\prod_{\theta_k\not\in A} \left[1-\exp(-w_k^-)\right]\right\}\left\{\prod_{\theta_k\in A} \exp(-w_k^-)\right\}
\end{equation*}
 for any $A\subseteq \Theta$ such that $|A|>1$, where $C$ is a proportionality constant.
\end{Prop}
\noindent \emph{Proof:} see  \ref{sec:mass}.

\section{Identification of model parameters}
\label{sec:ident}

To compute the output mass function given by Eq (\ref{eq:m12}) in the binary case and by Proposition \ref{prop:mass} in the multi-category case, we need to compute the weights of evidence. In the binary case, these weights depend on coefficients $\beta_j$ and $\alpha_j$ for $j=1,\ldots,J$ through (\ref{eq:weights2}). A learning procedure (such as likelihood maximization) gives us estimates  $\betah_j$ of $\beta_j$ for $j=0,\ldots,J$. Parameters $\alpha_j$ are not identifiable, but are linked to $\beta_0$  by Eq. (\ref{eq:constr_alpha2}). In the multi-category case, things are worse, because parameters $\beta_{jk}$ are  also not identifiable:  we can easily check that adding any constant vector $\bc=(c_0,\ldots,c_J)$ to each vector $\beta_k=(\beta_{0k},\ldots,\beta_{Jk})$  produces the same normalized plausibilities (\ref{eq:normpl}). Both parameters $\beta_{jk}$ and $\alpha_{jk}$ are, thus, underdetermined in that case.
 
To identify the model parameters, we propose to apply the Least Commitment Principle introduced in Section \ref{subsubsec:LCP}, by searching for the parameter values that give us the output mass functions with minimal information content,  the information content of a mass function $m$ being taken  to be  $I_p(m)$ defined by (\ref{eq:Ip}), with $p=2$. (The value $p=2$ is chosen because it lends itself to easy computation, as will be shown below). We will first deal with the binary case in Section \ref{subsec:ident2} and proceed with the multi-category case in Section \ref{subsec:identK}.  

\subsection{Binary case}
\label{subsec:ident2}

Let $\{(x_i,y_i)\}_{i=1}^n$ be the learning set, let $\betah_{j}$ be the maximum likelihood estimate of the coefficients $\beta_{j}$, and let $\balpha$ denote the vector  $(\alpha_{1},\ldots,\alpha_J)$. The  values  $\alpha_{j}^*$ minimizing the sum of the squared weights of evidence can thus be found by solving the following minimization problem
\begin{equation}
\label{eq:critere2}
\min f(\balpha)= \sum_{i=1}^n \sum_{j=1}^J \left(\betah_{j}\phi_j(x_i) + \alpha_{j}\right)^2
\end{equation}
subject to
\begin{equation}
\label{eq:constr2}
\sum_{j=1}^J \alpha_{j} = \betah_0.
\end{equation}
Developing the square in (\ref{eq:critere2}), we get
\begin{equation}
\label{eq:f2}
f(\balpha) =\sum_{j=1}^J  \betah_{j}^2 \left(\sum_{i=1}^n \phi_j(x_{i})^2 \right)+ n\sum_{j=1}^J \alpha_j^2  +
2 \sum_{j=1}^J \betah_j\alpha_j\sum_{i=1}^n \phi_j(x_i).  
\end{equation}
The first term in the right-hand side of (\ref{eq:f2}) does not depend on $\balpha$, and the third term vanishes when the $J$ features are centered, i.e., when  $\sum_{i=1}^n \phi_j(x_i)=0$ for all $j$. Let us first assume that this condition is met. Then, we just need to minimize $\sum_{j=1}^J \alpha_j^2$ subject to (\ref{eq:constr2}). The solution is 
\begin{equation}
\label{eq:alphastar}
\alpha_j^*=\betah_0/J, \quad j=1,\ldots,J.
\end{equation}
In the case of logistic regression, where $\phi_j(x)=x_j$, the condition $\sum_{i=1}^n \phi_j(x_i)=0$ can easily be ensured by centering the data before estimating the parameters. In the nonlinear case, the features $\phi_j$ are constructed during the learning process and they cannot be centered beforehand. Let $\mu_j$ denote the mean of feature $\phi_j$,
$
\mu_j =\frac{1}{n} \phi_j(x_i),
$
and $\phi'(x_i)=\phi(x_i)-\mu_j$ the centered feature values. We can write
\[
w_{ij}=\beta_j \phi_j(x_i)+ \alpha_j=\beta_j \phi'_j(x_i)+ \alpha'_j,
\]
with 
$
\alpha'_j=\alpha_j + \beta_j\mu_j,
$
and
\[
\sum_{j=1}^J w_{ij}= \sum_{j=1}^J\beta_j \phi_j(x_i)+\beta_0=\sum_{j=1}^J\beta_j \phi'_j(x_i)+\beta'_0
\]
with 
$
\beta'_0=\beta_0 +\sum_{j=1}^J \beta_j\mu_j.
$
As shown above, the optimal value of $\alpha'_j$ is
\[
\alpha_j'^*=\frac{\betah_0'}{J}=\frac{\betah_0}J + \frac1J \sum_{j=1}^J \betah_j\mu_j.
\]
Consequently, the optimal value of $\alpha_j$ is
\begin{equation*}
\alpha_j^*=\alpha_j'^*-\betah_j\mu_j=\frac{\betah_0}J + \frac1J \sum_{q=1}^J \betah_q\mu_q -\betah_j\mu_j.
\end{equation*}

\begin{Rem}
\label{rem:regularization}
In this section, we have started from  parameter estimates $\betah_j$, $j=0,\ldots,J$ to compute the values $\alpha_j^*$ that give us the least informative mass functions, in terms of the sum of  squared weights of evidence. We thus have a two-step process, where  coefficients $\beta_j$ are first estimated, and the $\alpha_j$ are determined in a second step. As a  complementary approach, we can attempt to minimize the squared weights of  evidence in the course of the learning process. In the simple case where the features are centered, the sum of squared weights of evidence has the following form, from (\ref{eq:f2}) and (\ref{eq:alphastar}):
\[
\sum_{j=1}^J \beta_j^2 \left(\sum_{i=1}^n \phi_j(x_{i})^2 \right)+ \frac{n}{J} \beta_0^2.
\]
As a heuristic, we can add to the loss function a term $\lambda_1 \sum_{j=1}^J \beta_j^2+\lambda_2 \beta_0^2$. We recognize the idea of  \emph{ridge regression} and $\ell_2$-regularization, or \emph{weight decay}. We can thus reinterpret regularization  in the last layer of a neural network as a heuristic for minimizing the sum of squared  weights of evidence, in application of the Least Commitment Principle. This  remark  also applies to the multi-category case addressed in the next section.
\end{Rem}

\subsection{Multi-category case}
\label{subsec:identK}

In the multi-category case, we must determine both sets of coefficients $\{\beta_{jk}\}$ and $\{\alpha_{jk}\}$. As before,  let $\betah_{jk}$ denote the maximum likelihood estimates of the weights $\beta_{jk}$, and let $\balpha$ denote the vector of parameters $\alpha_{jk}$. Any set of coefficients $\beta_{jk}^*=\betah_{jk}+c_j$ will produce the same output probabilities  (\ref{eq:normpl}) as $\betah_{jk}$. The optimal parameter values $\beta_{jk}^*$ and $\alpha_{jk}^*$  can, thus, be found by solving the following minimization problem
\begin{equation}
\label{eq:critere}
\min f(\bc,\balpha)= \sum_{i=1}^n \sum_{j=1}^J \sum_{k=1}^K  \left[(\betah_{jk}+c_j)\phi_j(x_{i}) + \alpha_{jk}\right]^2
\end{equation}
subject to the $K$ linear constraints
\begin{equation}
\label{eq:constr}
\sum_{j=1}^J \alpha_{jk} = \betah_{0k} + c_0, \quad k=1,\ldots,K.
\end{equation}
\begin{Prop}
\label{prop2}
The solution of the minimization problem (\ref{eq:critere})-(\ref{eq:constr}) is given by
\[
\beta_{0k}^*=\betah_{0k}- \frac{1}{K} \sum_{l=1}^K \betah_{0l}
\] 
and
\begin{equation}
\label{eq:alphanc}
\alpha_{jk}^* =\frac1J\left(\beta_{0k}^*+\sum_{j=1}^J \beta^*_{jk}\mu_j\right)-\beta_{jk}^*\mu_j.
\end{equation}
\end{Prop}
\noindent \emph{Proof:} See \ref{sec:prop2}.

To get the least committed mass function $m_i^*$ with minimum sum of squared weights of evidence and verifying (\ref{eq:normpl}), we thus need to center the rows of the $(J+1)\times K$ matrix $B=(\betah_{jk})$, set  $\alpha_{jk}^*$ according to  (\ref{eq:alphanc}), and compute the weights of evidence $w_k^-$ and $w_k^+$ from (\ref{eq:weights1}) and (\ref{eq:weights3}).

\section{Numerical experiments}
\label{sec:results}

In this section, we illustrate through examples some properties of the mass functions computed by GLR classifiers. We demonstrate their use to interpret the computations performed by such networks, and to quantify prediction uncertainty. We start with a binary classification problem and logistic regression in Section \ref{subsec:2class}. We then proceed with a multi-category dataset and a neural network model in Section \ref{subsec:multiclass}.

\subsection{Heart disease data}
\label{subsec:2class}

As an example of a real dataset, we considered the Heart Disease data\footnote{This dataset can be downloaded from \url{https://web.stanford.edu/~hastie/ElemStatLearn/}.} used in \cite{hastie09}. These data were collected as part of a study  aiming  to establish the intensity of ischemic heart disease risk factors in a high-incidence region in South Africa. The data represent white males between 15 and 64, and the response variable is the presence or absence of myocardial infarction (MI) at the time of the survey. There are 160 positive cases in this data, and a sample of 302 negative cases (controls). For display purposes, we considered only two input variables: age and low-density lipoprotein (LDL) cholesterol. The output variable $Y$ takes values $\theta_1$ and $\theta_2$ for presence and absence of MI, respectively.

\begin{figure}
\centering  
\includegraphics[width=0.5\columnwidth]{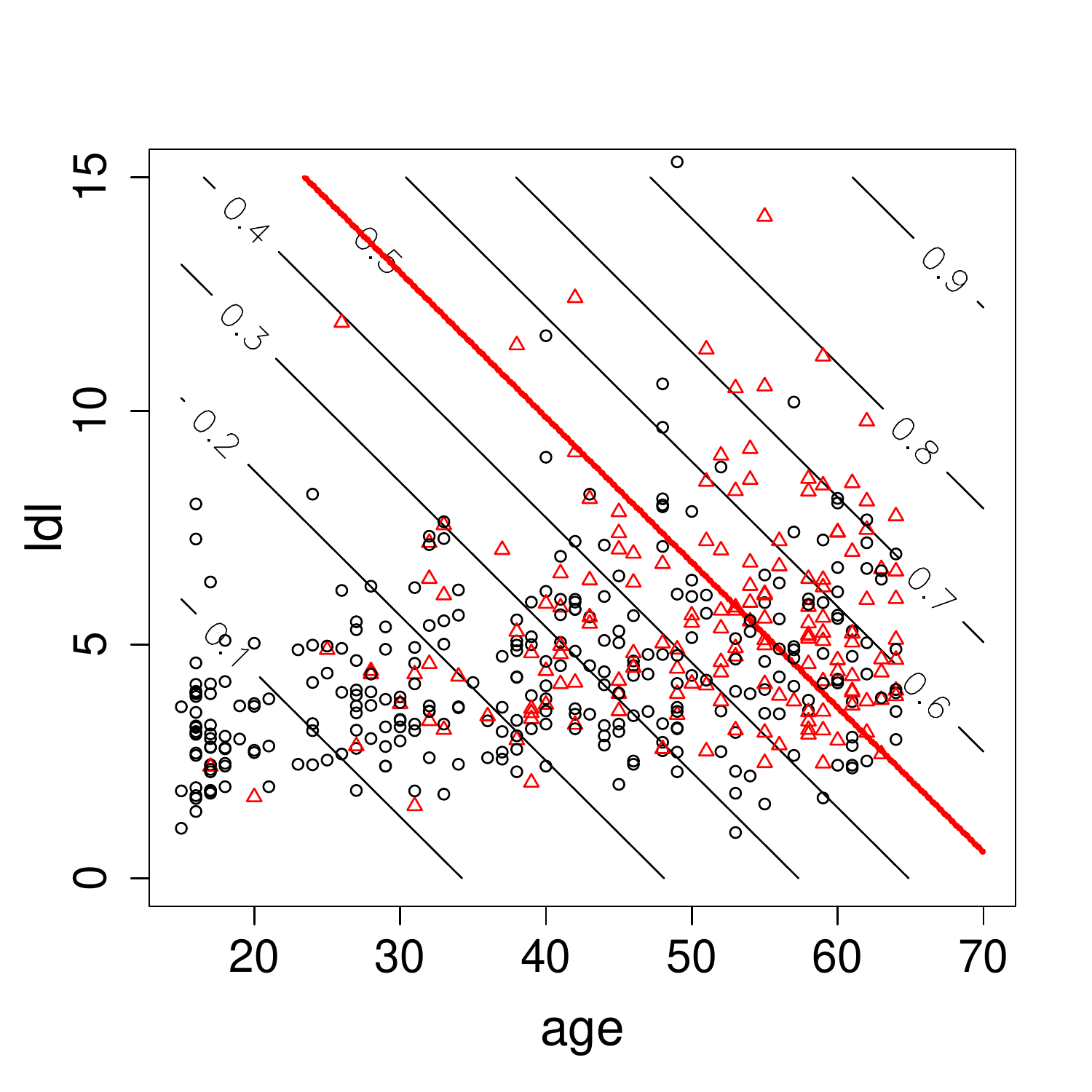}
\caption{Heart disease data, with the decision boundary (thick solid line) and the lines of equal positive class posterior probability for the logistic regression classifier. The positive and negative instances are identified by triangles and circles, respectively. \label{fig:heart_proba}}
\end{figure}

\subsubsection{Analysis and interpretation of mass functions}

\begin{figure}
\centering  
\subfloat{\includegraphics[width=0.4\columnwidth]{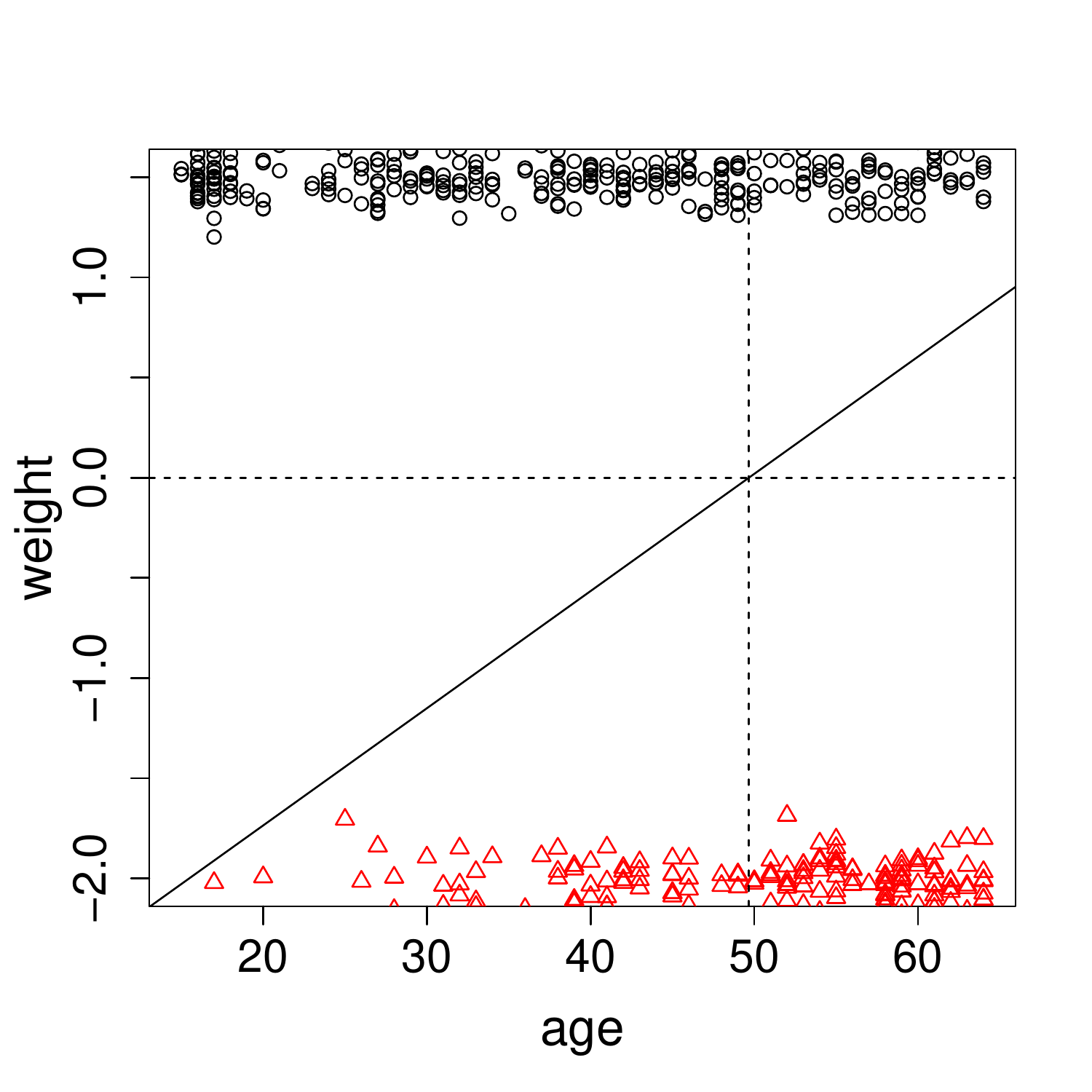}}
\subfloat{\includegraphics[width=0.4\columnwidth]{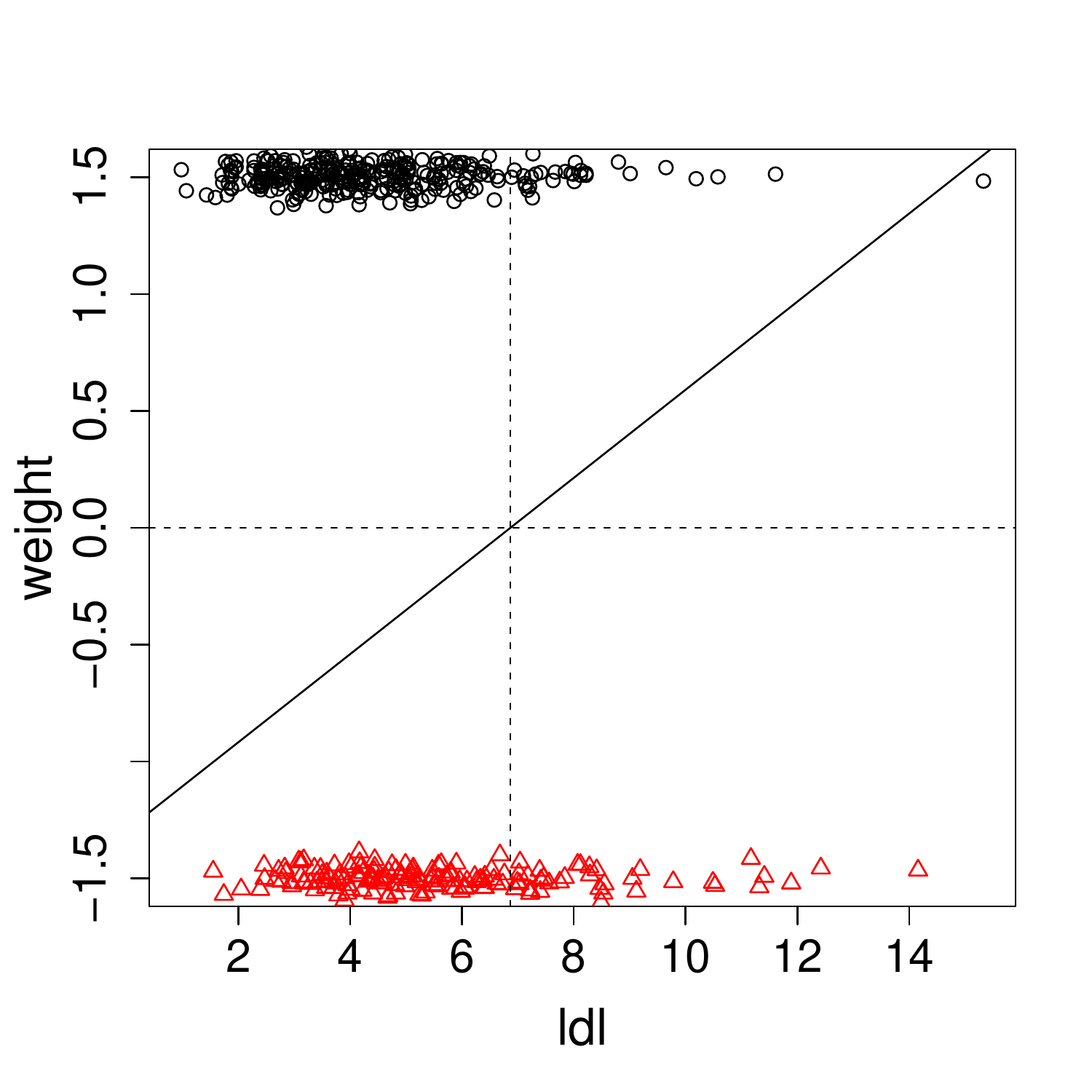}}
\caption{Weights $w_j$  as a function of $x_j$ for variables age  (left) and LDL (right). The feature values for positive and negative instances are shown, respectively, on the lower and upper  horizontal axes, with some random vertical jitter to avoid overlap. \label{fig:heart_weights}}
\end{figure}

\begin{figure}
\centering  
\subfloat{\includegraphics[width=0.4\columnwidth]{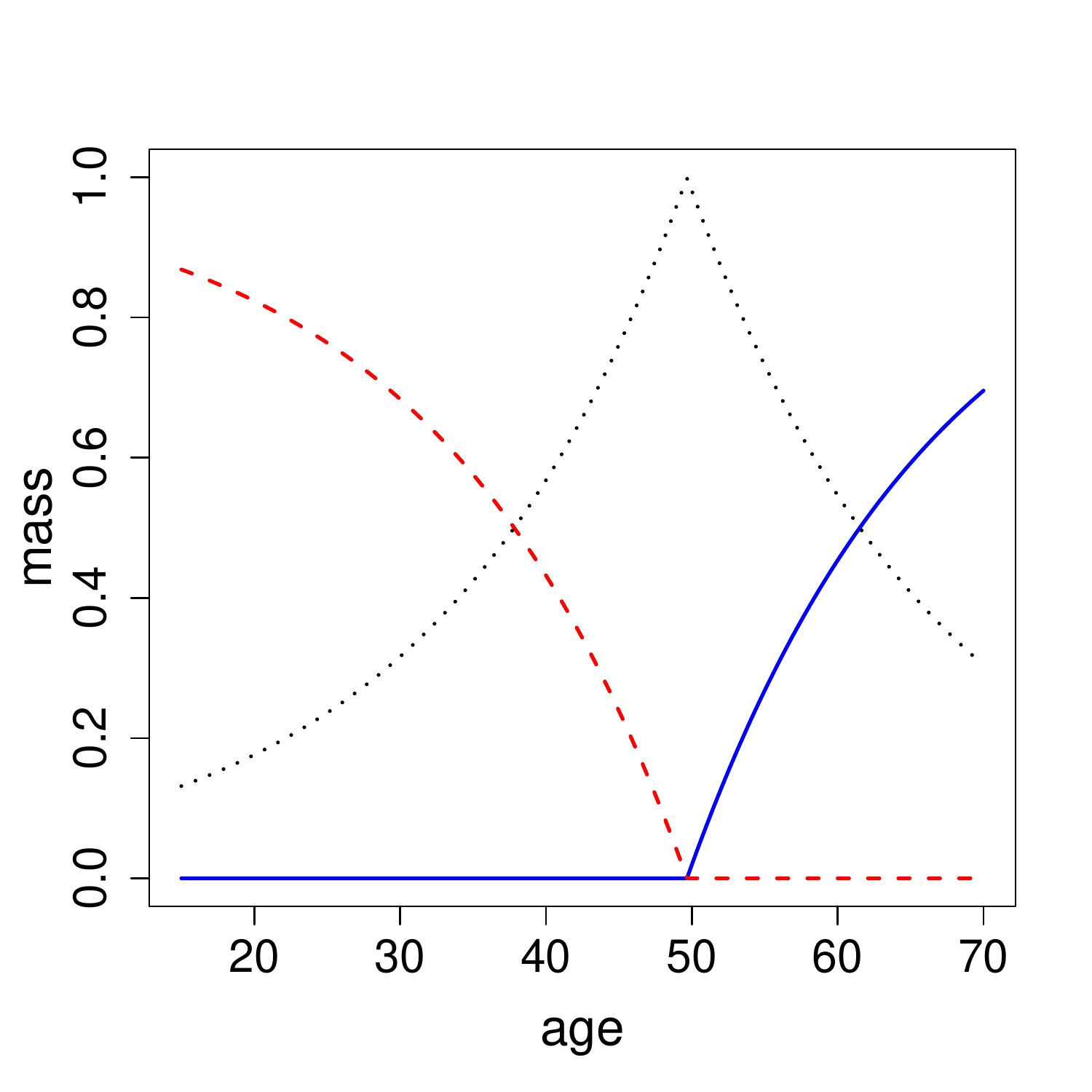}}
\subfloat{\includegraphics[width=0.4\columnwidth]{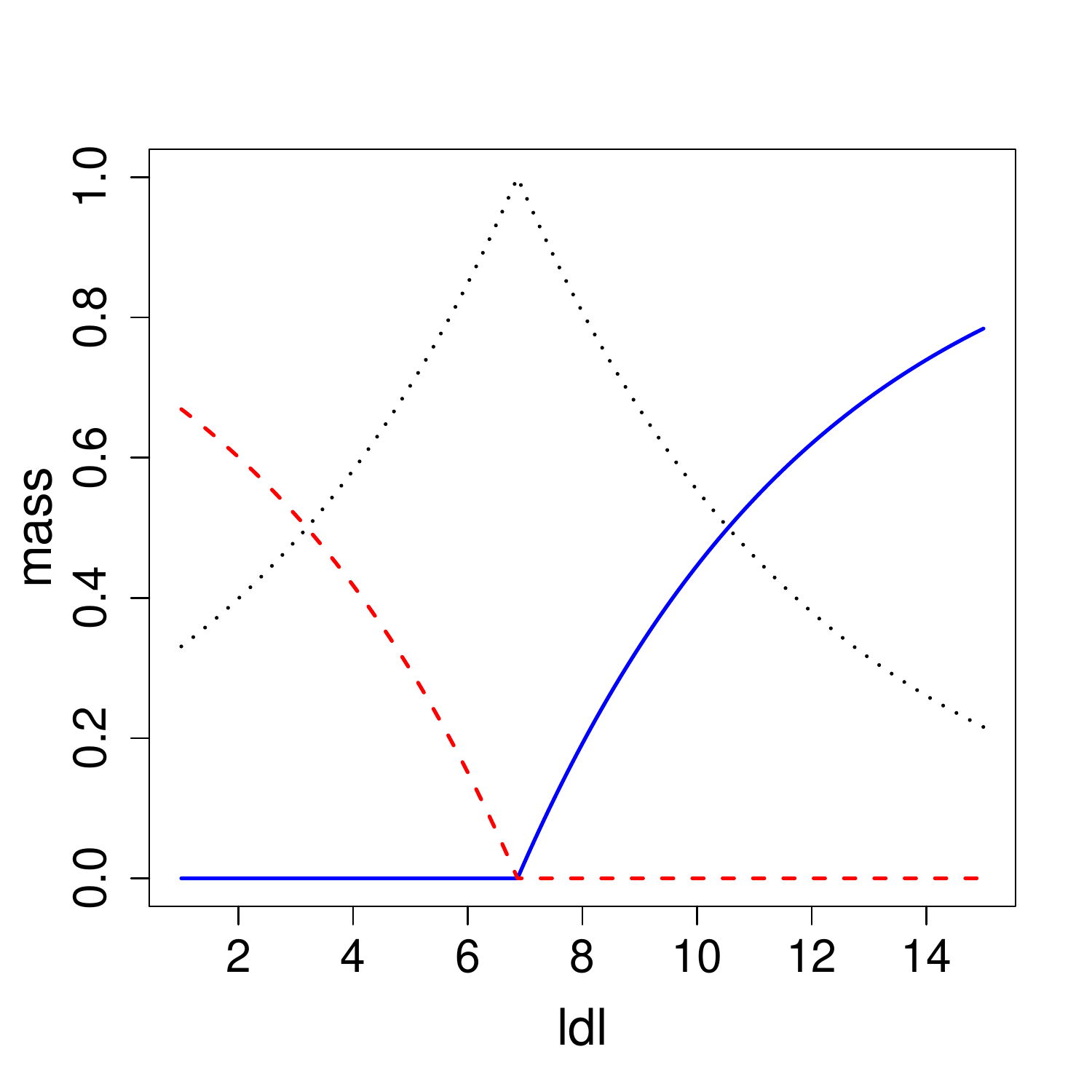}}
\caption{Mass functions $m_j$  for variables age  (left) and LDL (right). The solid, broken and dotted lines correspond, respectively, to $m_j(\stheta{1})$, $m_j(\stheta{2})$ and $m_j(\Theta)$. \label{fig:heart_masses}}
\end{figure}

Figure \ref{fig:heart_proba} shows the data, with the decision boundary and the lines of equal class $\theta_1$ posterior probability for the logistic regression classifier. The weights of evidence $w_j$ as functions of $x_j$ for the two input variables are shown in Figure \ref{fig:heart_weights}. We can see that an age greater than $\xi_1\approx50$  is evidence for the presence of MI ($w_j>0$), whereas an age less than 50 is evidence for the absence of MI ($w_j<0$). For LDL, the cut-off point is $\xi_2\approx 6.87$. The corresponding mass functions $m_j$ for each of the two features are displayed in Figure \ref{fig:heart_masses}. At the cut-off point  $\xi_j$, the mass function $m_j$ is vacuous, which indicates that feature $x_j$ does not support any of the two classes.

\begin{figure}
\centering  
\subfloat[\label{subfig:heart_belief}]{ \includegraphics[width=0.4\columnwidth]{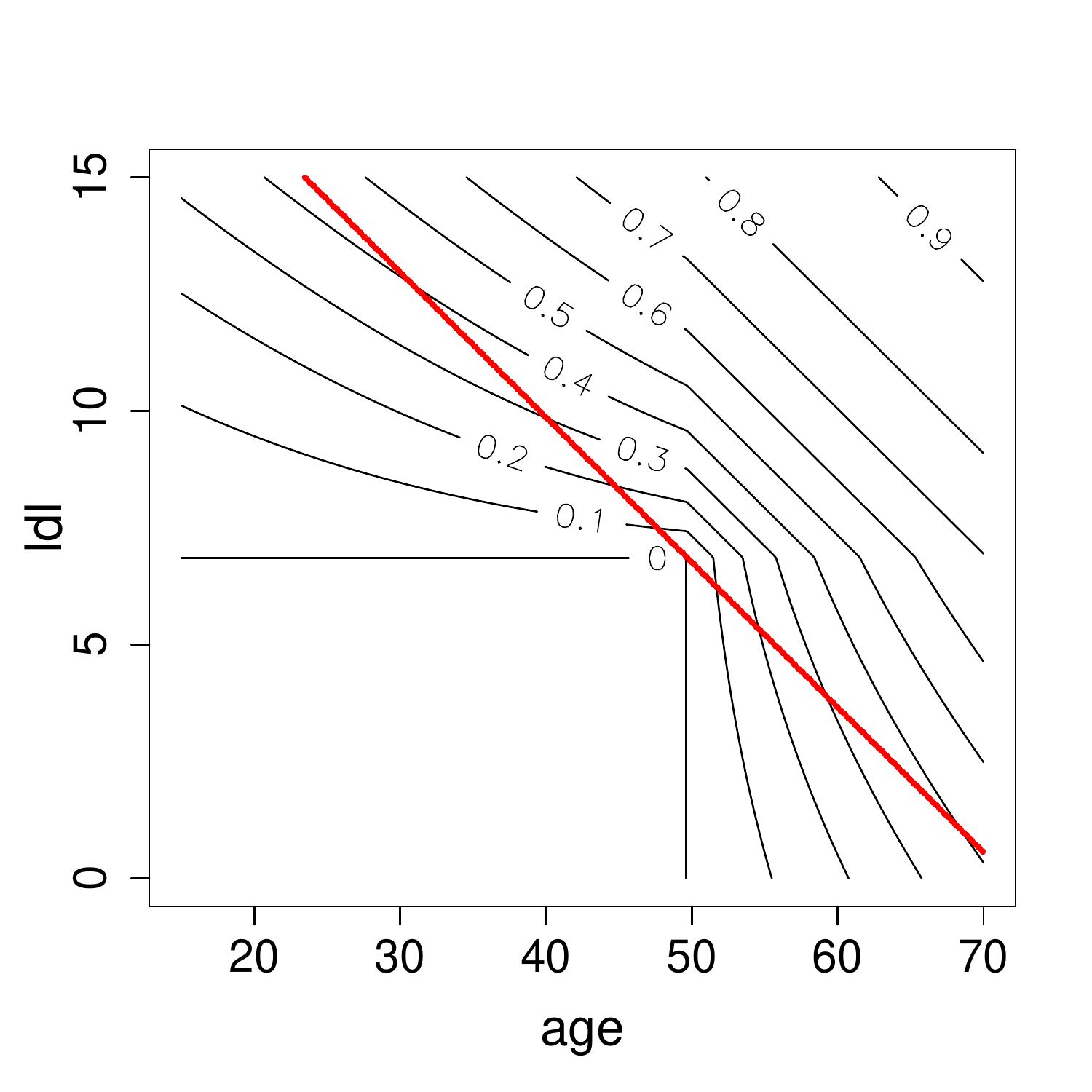}}
\subfloat[\label{subfig:heart_plausibility}]{\includegraphics[width=0.4\columnwidth]{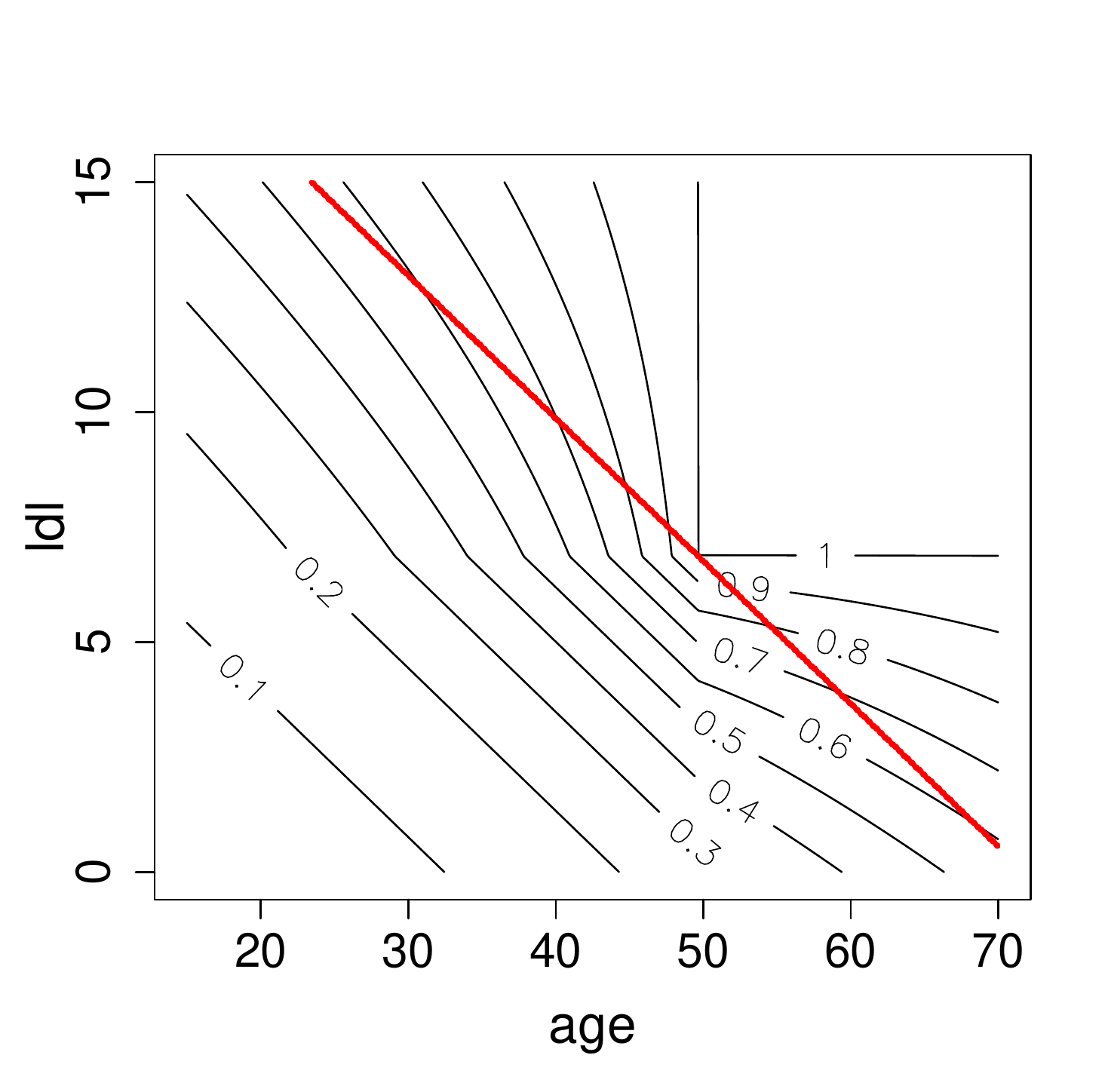}}\\
\subfloat[\label{subfig:heart_Thetaconflit}]{ \includegraphics[width=0.4\columnwidth]{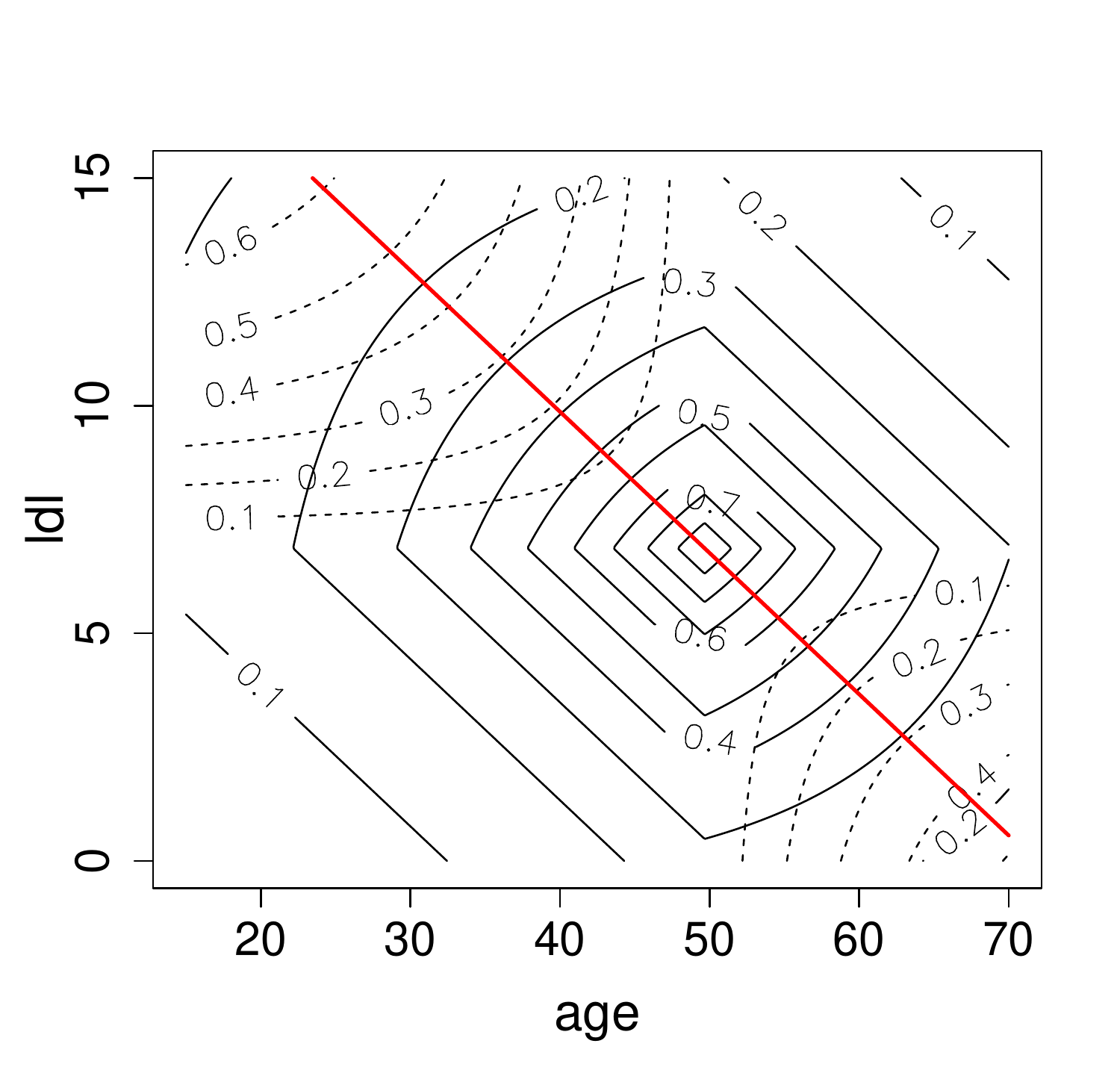}}
\subfloat[\label{subfig:heart_decision}]{\includegraphics[width=0.4\columnwidth]{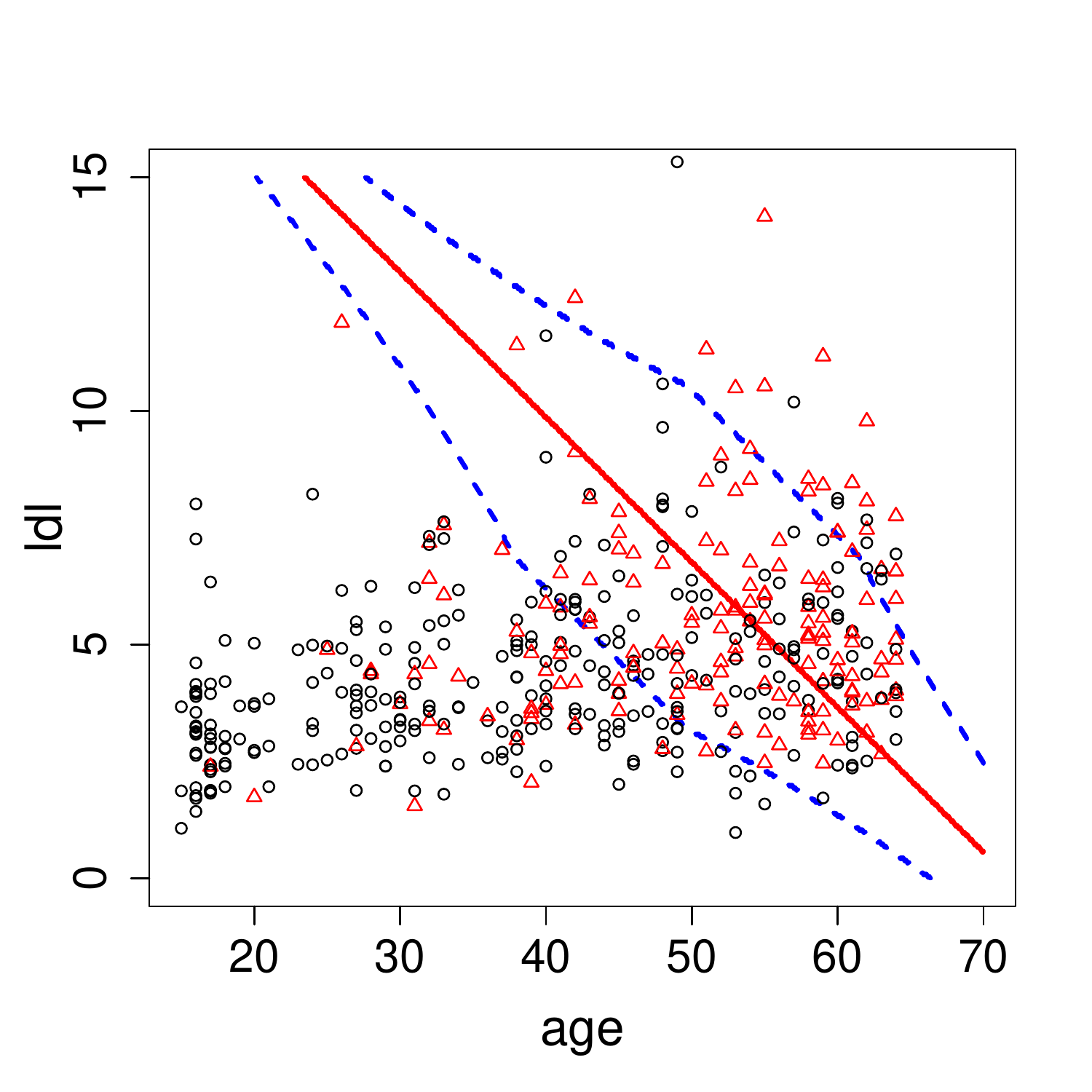}}
\caption{(a): Curves of equal degree of belief $Bel(\{\theta_1\})=m(\{\theta_1\})$ for the positive class; (b): curves of equal plausibility $pl(\theta_1)$; (c): ignorance $m(\Theta)$ (solide lines) and degree of conflict (broken lines); (d): Decision boundaries for the MP rule (solid line) and the ID rule (broken lines). \label{fig:heart_2D}}
\end{figure}

Different views of the output mass functions $m$ obtained after combining the two feature-based mass functions $m_j$, $j=1,2$ are shown in Figure \ref{fig:heart_2D}. We can see that there is no support for the positive class when both variables are below their cut-off points (lower-left part of Figure \ref{subfig:heart_belief}), whereas the positive class is fully plausible (i.e., there is no support for the negative class) when both variables are above their cut-off points (upper-right Figure \ref{subfig:heart_plausibility}). When both variables are close to their cut-off points, the ignorance $m(\Theta)$ is high (Figure \ref{subfig:heart_Thetaconflit}). The conflict between the two feature mass functions $m_1$ and $m_2$ is high when the two pieces of evidence point two different hypotheses as it is the case, for instance, for a young subject with a high LDL level (upper-left  corner of Figure \ref{subfig:heart_Thetaconflit}). We can see that the DS perspective allows us to distinguish between  lack of support, and conflicting evidence. In the classic probabilistic setting, both cases result in posteriori probabilities close to 0.5, as shown in Figure \ref{fig:heart_proba}. Information about the nature of the evidence that gave rise to the posterior class probabilities is lost when normalizing the contour function.

\subsubsection{Decision analysis}

With 0-1 losses, the pessimistic (maximum belief) and optimistic (MP) decisions rules based on output mass functions yield the same results as the decision rule based on output probabilities because, from Eqs. (\ref{eq:m12}) and (\ref{eq:pltheta2}),
\[
p(\theta_1)\ge p(\theta_2) \Leftrightarrow Bel(\{\theta_1\})\ge Bel(\{\theta_2\})\Leftrightarrow Pl(\{\theta_1\})\ge Pl(\{\theta_2\}).
\]
The corresponding decision boundary is shown as a solid line in Figure \ref{subfig:heart_decision}. In contrast, the ID rule (recalled in Section \ref{subsubsec:decision}) leads to the decision regions delimited by broken lines in Figure \ref{subfig:heart_decision}. In the central region between the two curves, the intervals $[Bel(\{\theta_1\}),Pl(\{\theta_1\})]$ and $[Bel(\{\theta_2\}),Pl(\{\theta_2\})]$ are overlapping: consequently, the decision is $\{\theta_1,\theta_2\}$, i.e., there is not enough evidence to support selecting any of the two classes. Tables \ref{tab:conf_plaus} and \ref {tab:conf_ID} show, respectively, the confusion matrices for the MP and ID rules, estimated by 10-fold cross-validation. (The results shown are averages over 30 replications of 10-fold cross-validation). The numbers in Tables \ref{tab:conf_plaus} and \ref {tab:conf_ID} are expressed in percent and sum to 100. For instance, we can see from Table \ref{tab:conf_plaus} that, on average, 13\% of the data were in the positive class and were correctly classified by the MP rule, while 10.6\% of the data were in the negative class and were wrongly classified by the same rule. The MP rule had an estimated error rate of  $20.8+10.6=31.4\%$, while the ID rule had an error rate of $2.2+9.8=12.0\%$ and a rejection rate of 42.2\%. If the rejected instances were classified randomly, the mean error rate would be $12+42.2/2= 33.1\%$, which is only slightly higher than the MP error rate. This means that the ID rule is not overly cautious: it rejects instances that could hardly be classified better than randomly.


\begin{table}
\caption{Confusion matrix for the MP rule, in \%  (Heart data). \label{tab:conf_plaus}}
\centering
\begin{tabular}{l|l|c|c|}
\multicolumn{2}{c}{}&\multicolumn{2}{c}{True class}\\
\cline{3-4}
\multicolumn{2}{c|}{}&Positive ($\theta_1$)&Negative ($\theta_2$)\\
\cline{2-4}
\multirow{2}{*}{Predicted}& Positive ($\theta_1$)& 13.8  & 10.6 \\
\cline{2-4}
& Negative ($\theta_2$)& 20.8 & 54.8 \\
\cline{2-4}
\end{tabular}
\end{table}
   

\begin{table}
\caption{Confusion matrix for the ID rule, in \% (Heart data). \label{tab:conf_ID}}
\centering
\begin{tabular}{l|l|c|c|}
\multicolumn{2}{c}{}&\multicolumn{2}{c}{True class}\\
\cline{3-4}
\multicolumn{2}{c|}{}&Positive ($\theta_1$)&Negative ($\theta_2$)\\
\cline{2-4}
& Positive ($\theta_1$)& 3.6 & 2.2 \\
\cline{2-4}
Predicted & Negative ($\theta_2$)& 9.8 & 42.2 \\
\cline{2-4}
& $\{\theta_1,\theta_2\}$  &21.2 &  21.0
 \\
\cline{2-4}
\end{tabular}
\end{table}

\subsection{Gaussian Multi-category  Data}
\label{subsec:multiclass}

As an example of a multi-category classification task with nonlinear decision boundaries, we consider an artificial dataset with $d=2$ features, $K=3$ equiprobable classes, and Gaussian class-conditional densities: $X\vert Y=\theta_k \sim \calN(\mu_k,\Sigma_k)$, with
\[
\mu_1=\mu_2=(0,0)^T, \quad \mu_3=(1,-1)^T
\]
\[
\Sigma_1=0.1 I, \quad \Sigma_2=0.5 I, \quad \Sigma_3=\begin{pmatrix} 0.3 & -0.15\\-0.15 & 0.3\end{pmatrix},
\]
where $I$ is the $2\times 2$ identity matrix. We generated a learning set of size $n=900$, and we trained a neural network with two layers of 20 and 10 rectified linear units (ReLU) \cite{goodfellow16}. The output layer had a softmax activation function. The network was trained in batch mode with a mini-batch size of 100. The first hidden layer had a drop-out rate \cite{goodfellow16}  in the first hidden layer fixed to the standard value  of 0.5. The weights between the last hidden layer and the output layer were penalized with an $L_2$ regularizer and a coefficient $\lambda=0.5$ determined by 10-fold cross-validation.

\subsubsection{Mass functions}

\begin{figure}
\centering  
\includegraphics[width=0.5\columnwidth]{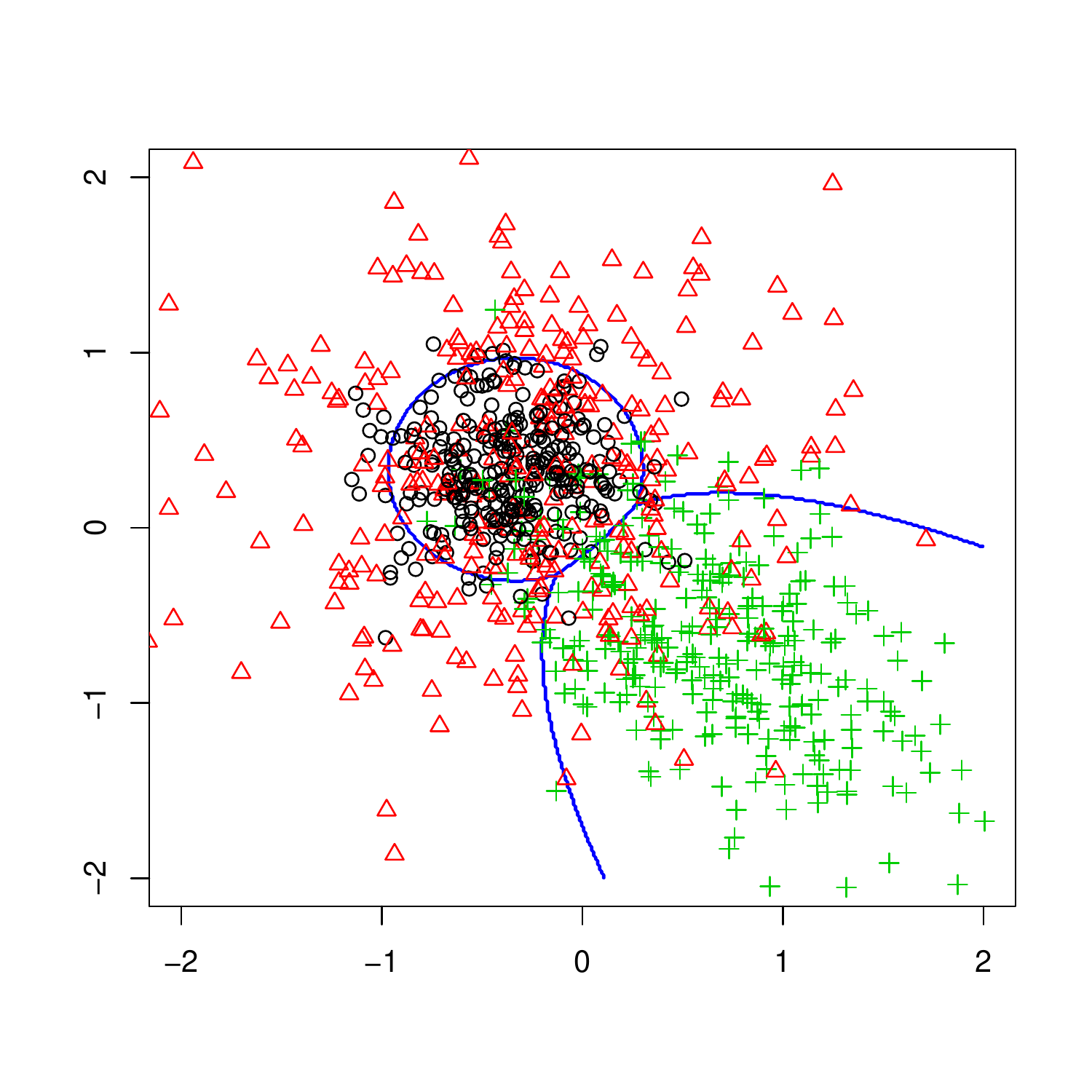}
\caption{Simulated Gaussian data with the Bayes decision boundary. \label{fig:data_non_linear}}
\end{figure}

\begin{figure}
\centering  
\subfloat{ \includegraphics[width=0.4\columnwidth]{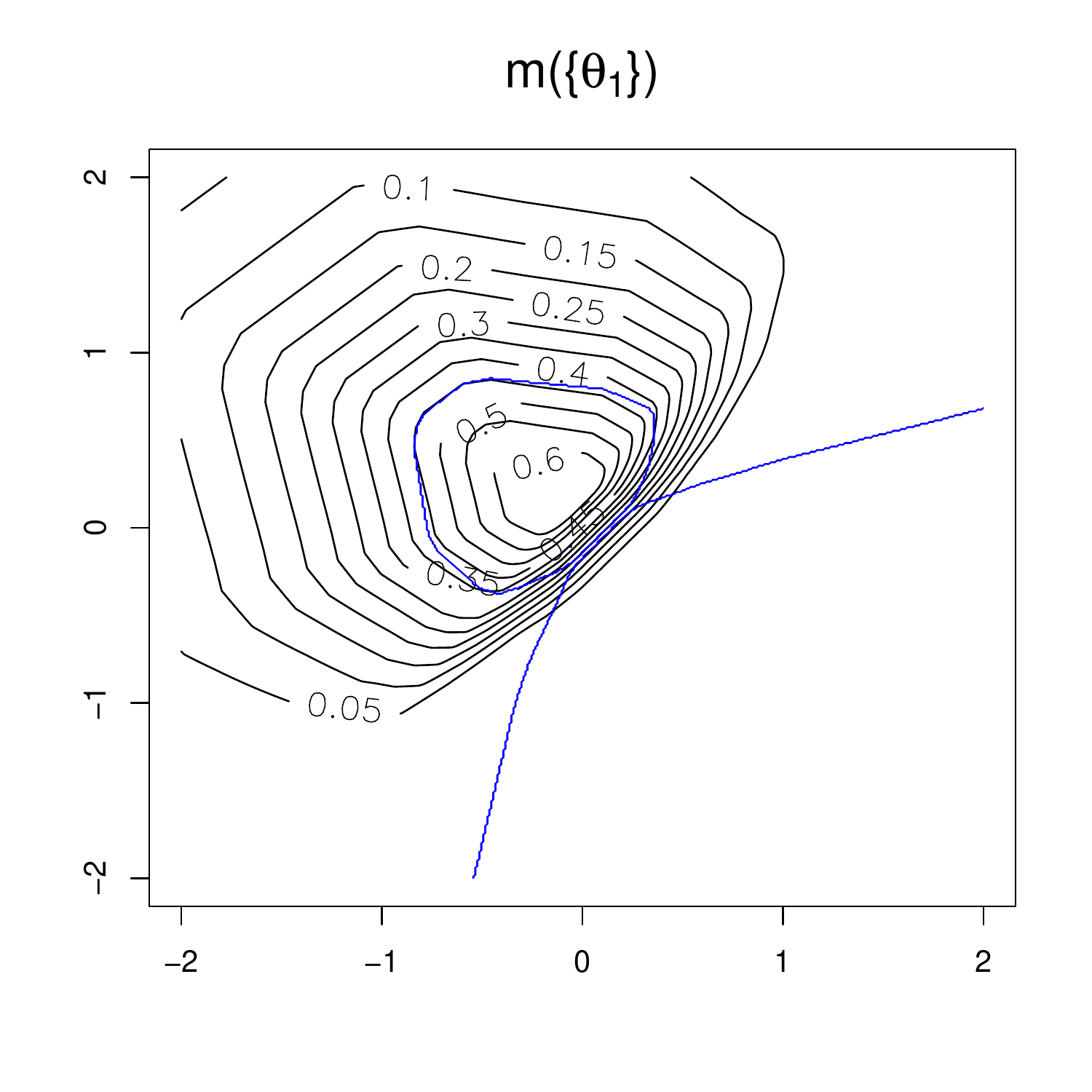}}
\subfloat{\includegraphics[width=0.4\columnwidth]{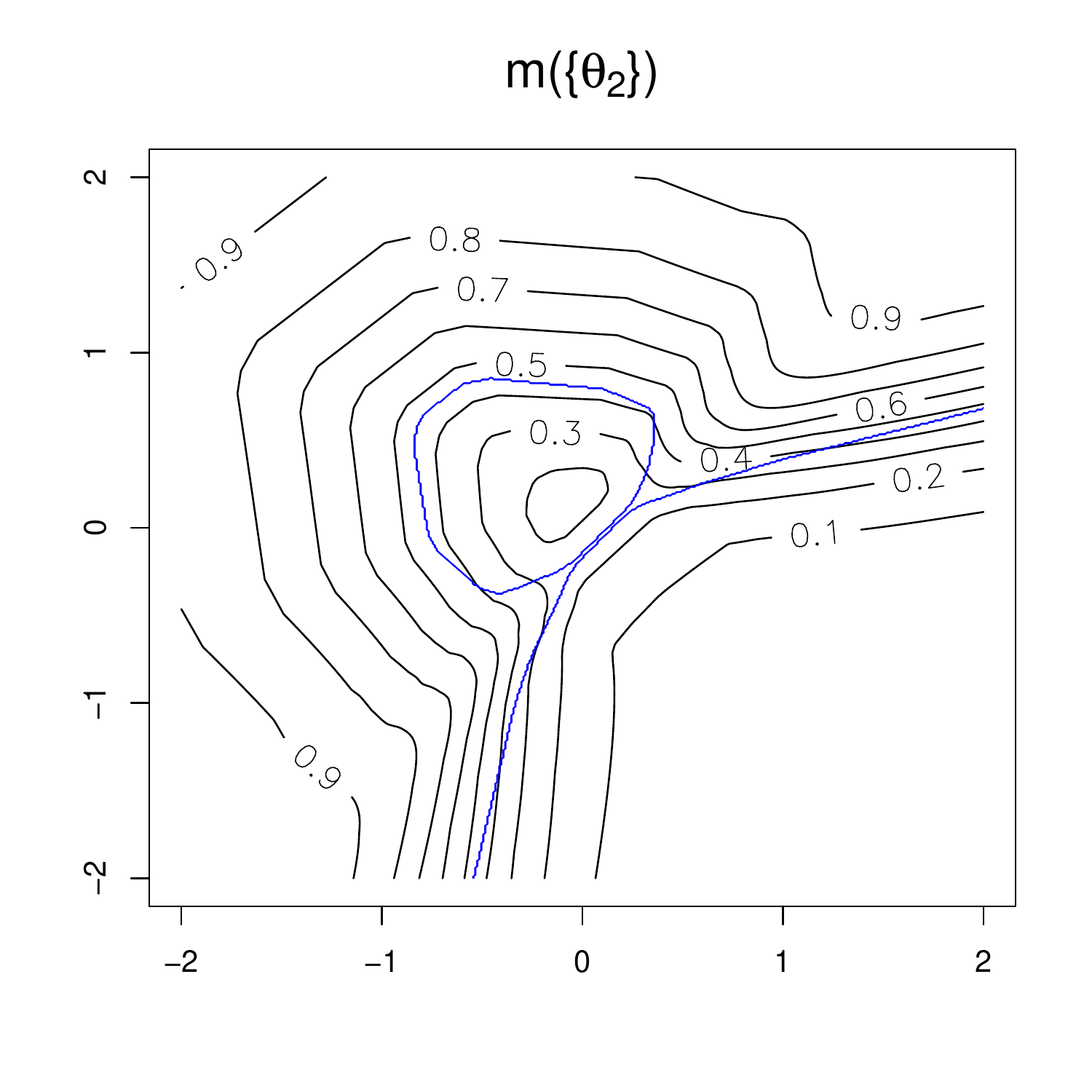}}\\
\subfloat{ \includegraphics[width=0.4\columnwidth]{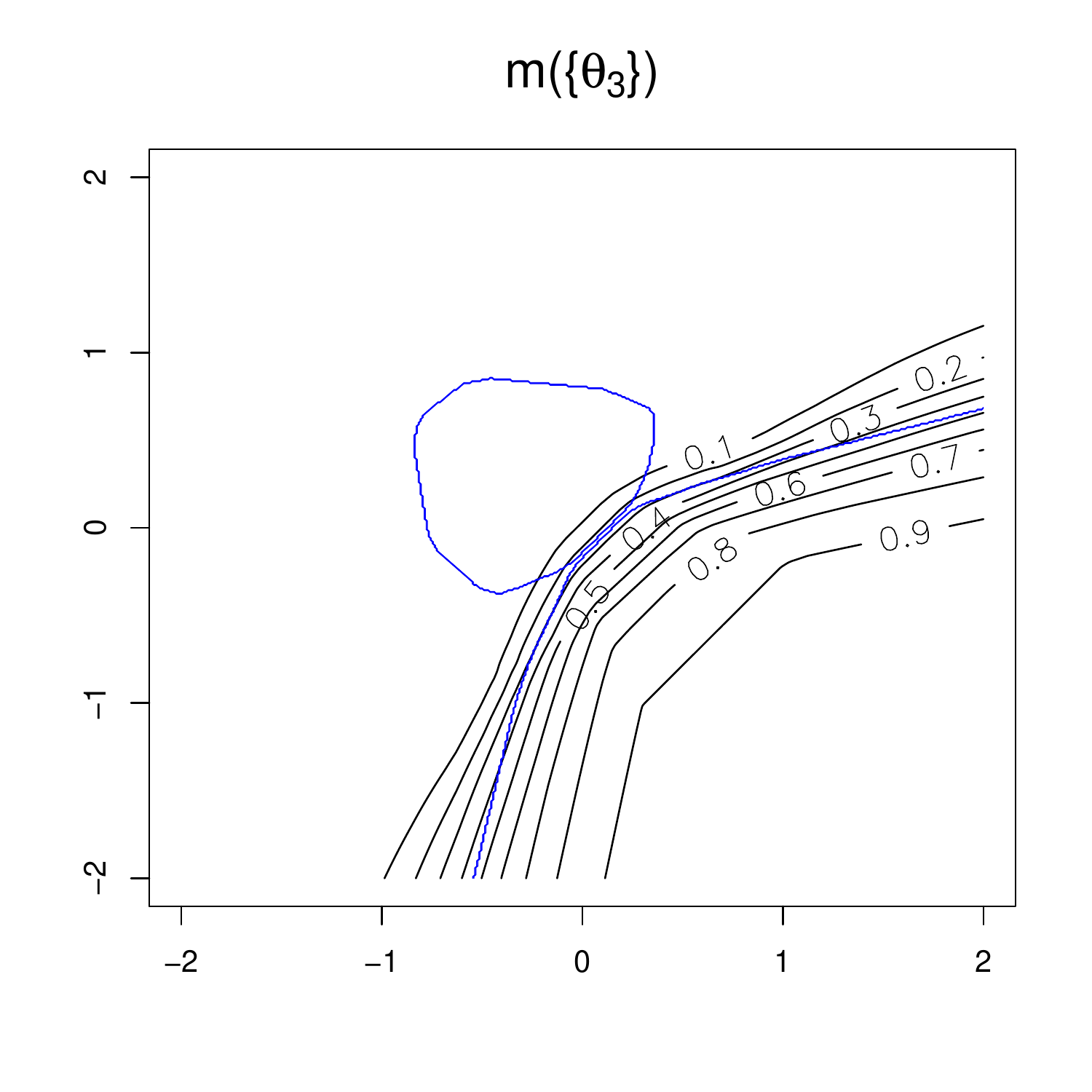}}
\subfloat{\includegraphics[width=0.4\columnwidth]{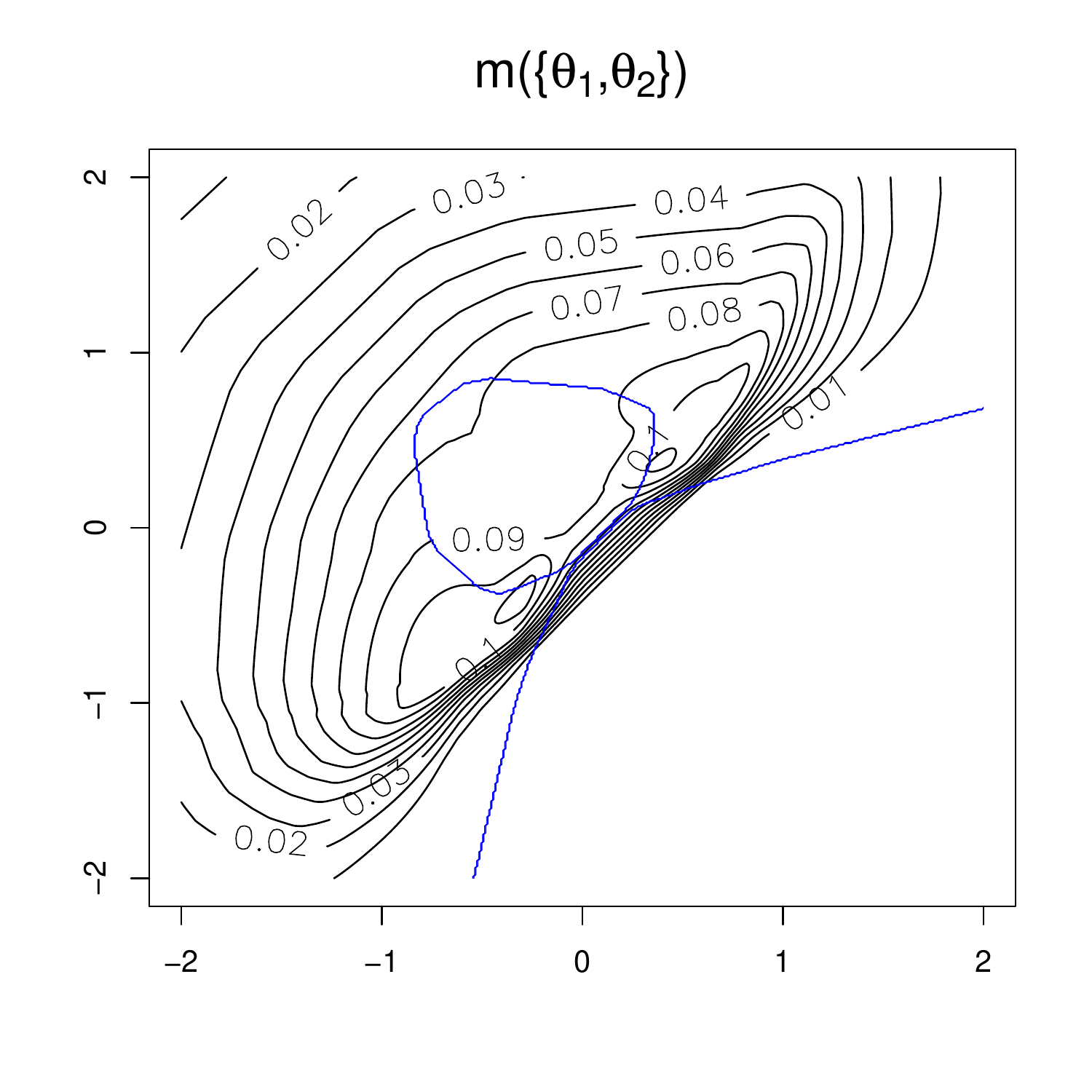}}\\
\subfloat{\includegraphics[width=0.4\columnwidth]{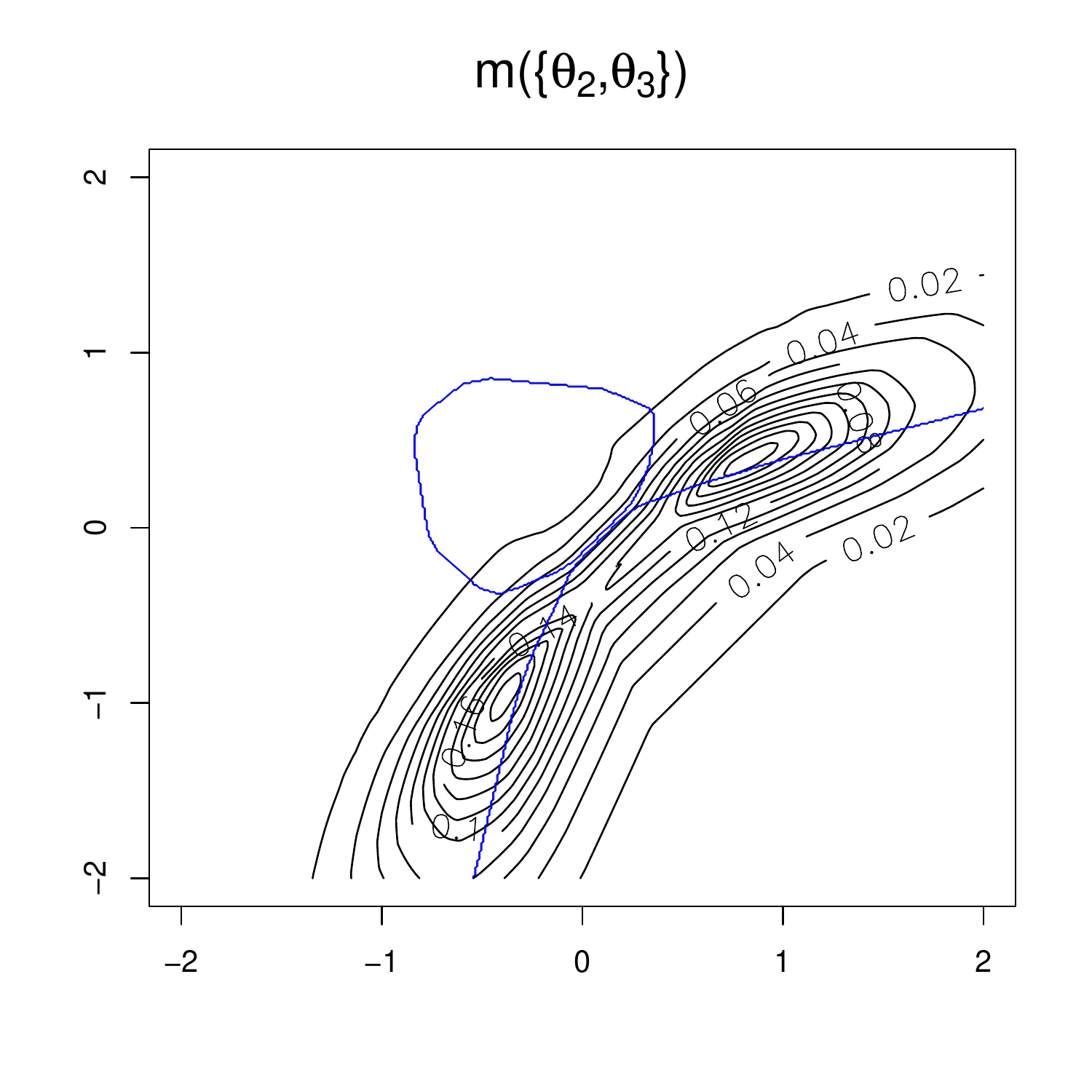}}
\subfloat{\includegraphics[width=0.4\columnwidth]{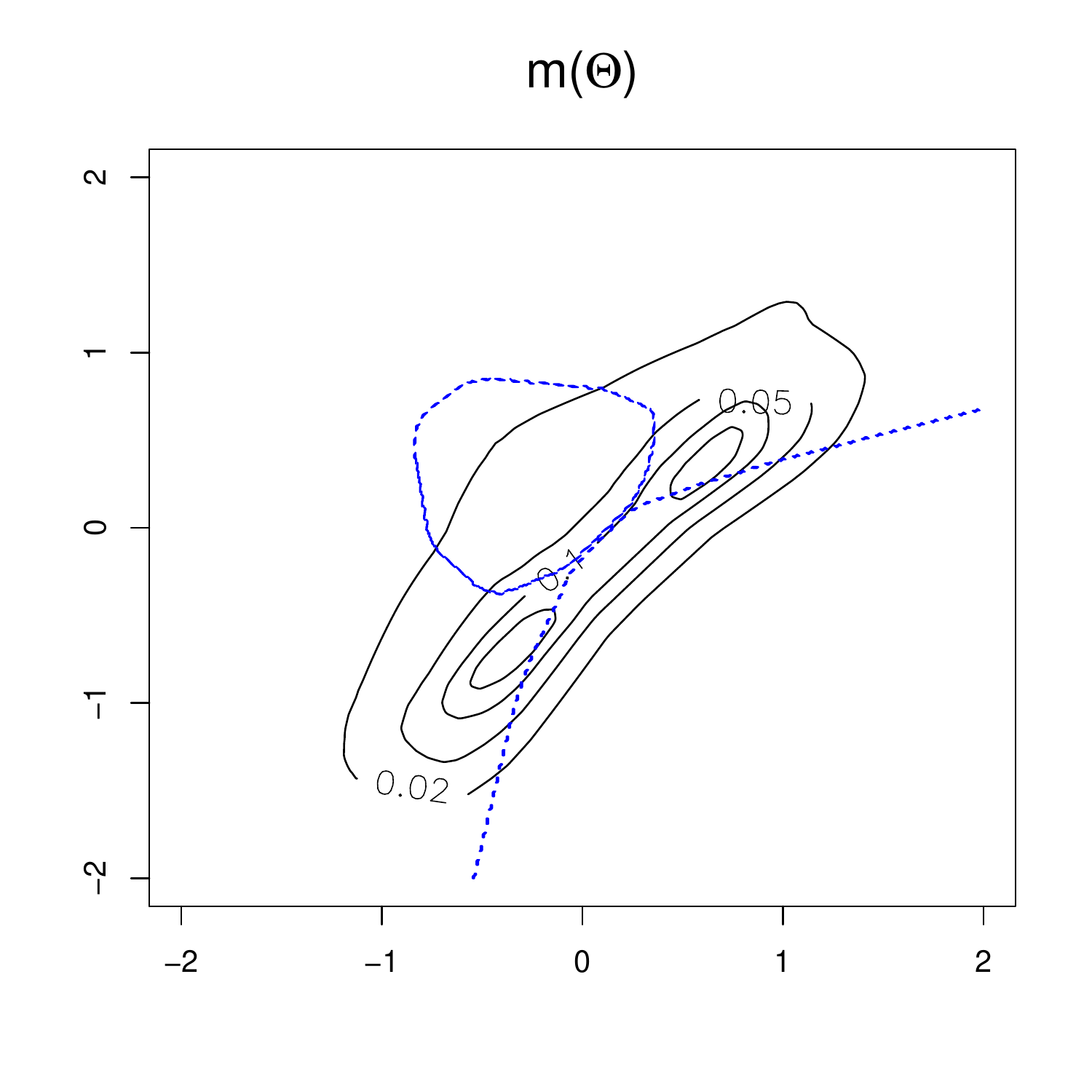}}
\caption{Level curves of the output masses  assigned to different focal sets (Gaussian data), with the MP decision boundary. \label{fig:NL_masses}}
\end{figure}

\begin{figure}
\centering  
\includegraphics[width=0.5\columnwidth]{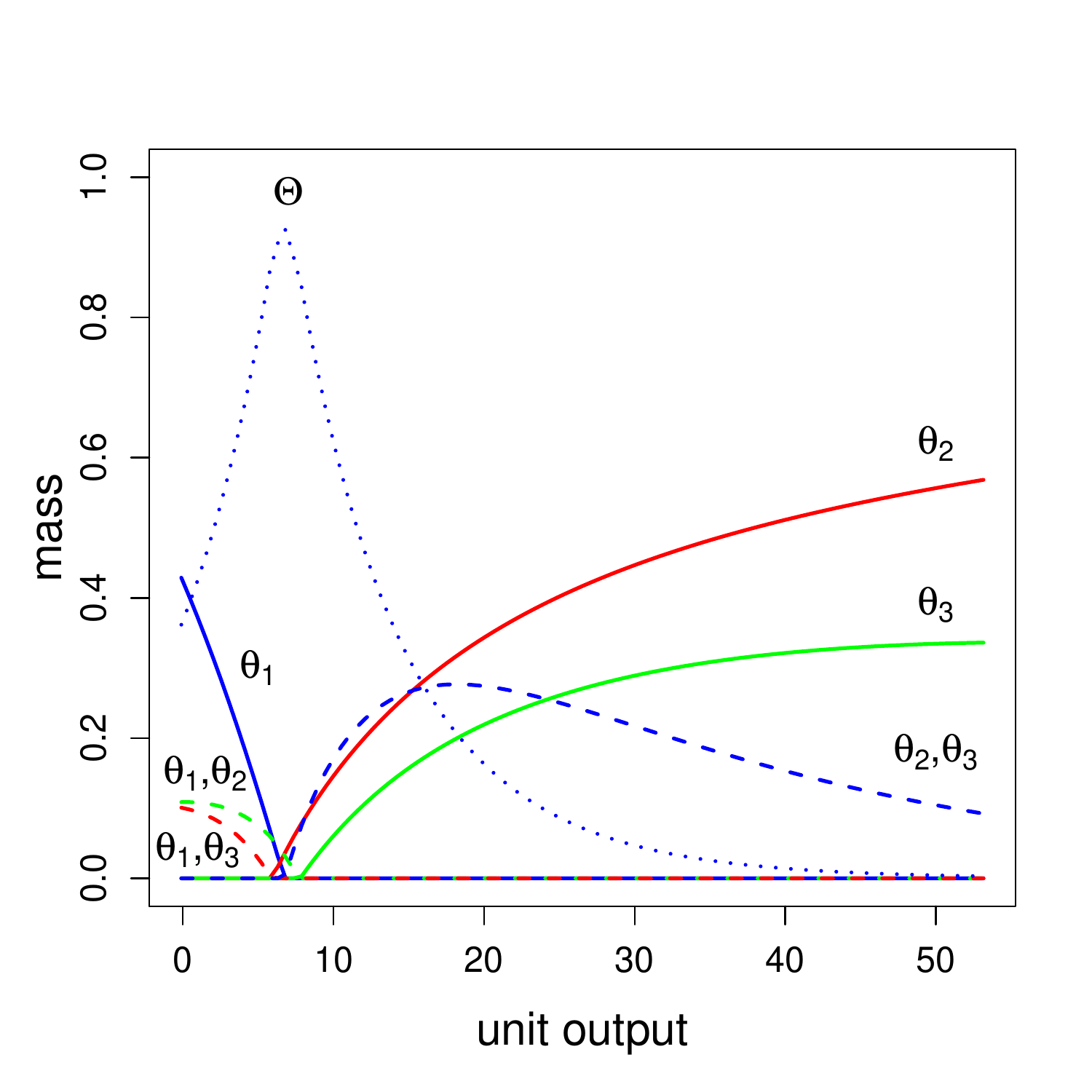}
\caption{Masses computed by a hidden unit, as functions of the unit outputs.  \label{fig:NL_unit2}}
\end{figure}

Figure \ref{fig:data_non_linear} shows the data and the Bayes decision boundary. Contour lines of the masses assigned to different focal sets are shown in Figure \ref{fig:NL_masses}. We can see that masses are assigned to singletons in region of high class density, and to sets of classes in regions where these classes overlap. The output mass function $m$ is the orthogonal sum of mass functions $m_j$ provided by the 10 units in the last hidden layer. Plotting these mass functions allows us to interpret the role played by  each of the hidden units. For instance, Figure \ref{fig:NL_unit2} shows the masses $m_j(A)$ assigned to different focal sets $A\subseteq \Theta$ by one of the hidden units. When the hidden unit output $\phi_j$ is small, the mass is distributed between $\{\theta_1\}$, $\{\theta_1,\theta_2\}$ and $\{\theta_1,\theta_3\}$. When $\phi_j$ is large, it supports $\{\theta_2\}$, $\{\theta_3\}$ and $\{\theta_2,\theta_3\}$.

\subsubsection{Decision boundaries}

\begin{figure}
\centering  
\includegraphics[width=0.5\columnwidth]{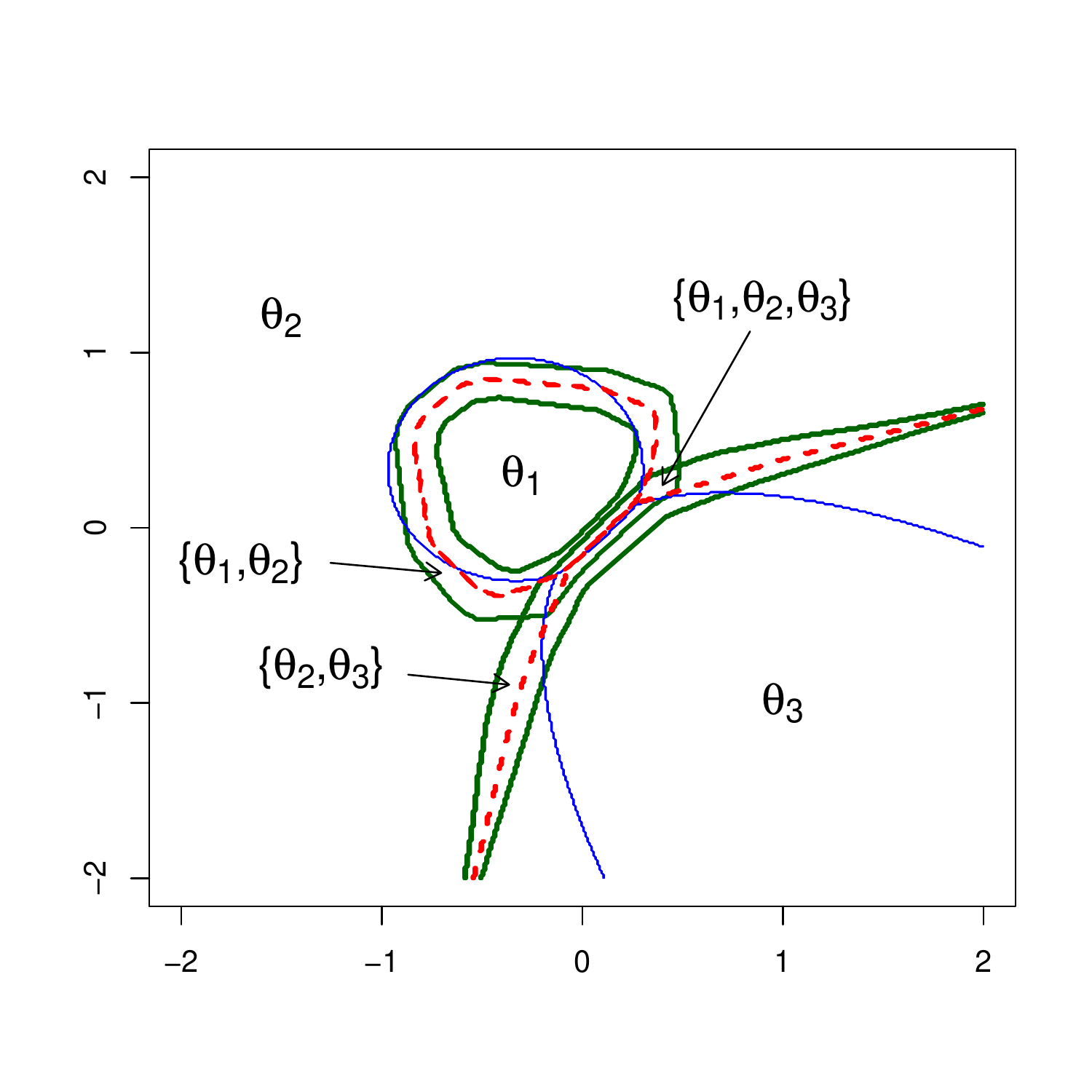}
\caption{Bayes decision boundary (thin solid line) and boundaries of the MP rule (thick broken line) and ID rule (thick solid line). The ID rule divides the feature space into six regions, with corresponding interpretations shown in the figure. \label{fig:regions}}
\end{figure}

The decision boundaries for the MP and ID rules are displayed in Figure \ref{fig:regions}. We can see that the ID rule divides the feature space into six decision regions, corresponding to precise assignment to each of the three classes, and to imprecise assignment to subsets $\{\theta_1,\theta_2\}$, $\{\theta_2,\theta_3\}$ and $\Theta=\{\theta_1, \theta_2,\theta_3\}$. The existence of these ``ambiguous'' decisions is due to lack of evidence in regions where the classes overlap. We observe that regions corresponding to sets of classes partially include the Bayes boundary: the Bayes optimal decision, thus, often belongs to the set of decisions prescribed by the ID rule, including cases where the MP rule differs from the Bayes decision.

\subsubsection{Error rates}

To estimate error rates, we generated a test dataset of size $n_t=15,000$. The estimated Bayes error rate was 24.6\%, and the estimated error rate of the MP rule was 25.7\%. The confusion matrices for the MP and ID rules are shown, respectively, in Tables \ref{tab:conf_plaus_gauss} and \ref{tab:conf_ID_gauss}. The error rate of the ID rule is 17.5\%, less that the Bayes error rate. Of course, this is compensated by assigning 16.46\% of instances to a pair of classes, and $2.83\%$ to the set of three classes. If one choses a single class in each decision set randomly, the mean error rate will be $17.5 + 16.46/2 + 2.83/3 \approx 26.7\%$, which is only slightly higher than the MP error rate. This result suggests that the neural network classifier indeed does not perform much better than chance when the ID rule does not select a single class.


 \begin{table}
\caption{Confusion matrix for the MP rule, in \% (Gaussian data). \label{tab:conf_plaus_gauss}}
\centering
\begin{tabular}{l|l|c|c|c|}
\multicolumn{2}{c}{}&\multicolumn{3}{c}{True class}\\
\cline{3-5}
\multicolumn{2}{c|}{}&$\theta_1$& $\theta_2$ & $\theta_3$\\
\cline{2-5}
& $\theta_1$& 26.8 & 9.7  &2.6 \\
\cline{2-5}
Predicted & $\theta_2$& 5.6 &  18.2 &  1.5 \\
\cline{2-5}
& $\theta_3$& 0.9  &5.4& 29.2\\
\cline{2-5}
\end{tabular}
\end{table}
    

\begin{table}
\caption{Confusion matrix for the ID rule, in \% (Gaussian data). \label{tab:conf_ID_gauss}}
\centering
\begin{tabular}{l|c|c|c|c|}
\multicolumn{2}{c}{}&\multicolumn{3}{c}{True class}\\
\cline{3-5}
\multicolumn{2}{c|}{}&$\theta_1$& $\theta_2$ & $\theta_3$\\
\cline{2-5}
& $\theta_1$& 21.5 & 6.6  &1.8 \\
\cline{2-5}
 & $\theta_2$& 2.5 &14.1  &0.5 \\
\cline{2-5}
Predicted & $\theta_3$&  0.9 & 5.2 &28.3\\
\cline{2-5}
& $\{\theta_1,\theta_2\}$&  6.8 & 5.3 & 1.0 \\
\cline{2-5}
& $\{\theta_2,\theta_3\}$& 0.3 & 1.4 & 1.2\\
\cline{2-5}
& $\{\theta_1,\theta_2,\theta_3\}$& 0.9 & 0.8 & 1.0 \\
\cline{2-5}
\end{tabular}
\end{table}

\section{Conclusion}
\label{sec:concl}

In this paper, we have revisited logistic regression and its extensions, including multilayer feedforward neural networks, by showing that these classifiers can be seen as converting (input or higher-level) features into mass functions and aggregating them by Dempster's rule of combination. The probabilistic outputs of these classifiers are the normalized plausibilities corresponding to the underlying combined mass function. This mass function has more degrees of freedom that the output probability distribution, and we have shown that it carries useful information. In particular, it makes it possible to distinguish between lack of evidence (when none of the features provides discriminant information) from conflicting evidence (when different features support different classes). This expressivity of mass functions allows us to gain insight into the role played by each input feature in  logistic regression, and to interpret hidden unit outputs in multilayer neural networks. It also makes it possible to use decision rules, such as the interval dominance rule, which select a set of classes when the available evidence does not unambiguously point to a single class, thus reducing the error rate.

The significance of this result stems, in our view, from the fact that it sets Dempster-Shafer theory as a suitable framework for analyzing and designing a wide range classifiers, including the now popular deep neural networks. Even though a lot of work has been done over the years applying belief functions to classification, this approach remained marginal in the vast landscape of statistical pattern recognition and machine learning techniques.  The results presented in this paper show that belief functions are, in fact, ubiquitous in a large number of machine learning algorithms, although this fact has been completely overlooked so far. This change of perspective opens the way to a whole research program, whose general objective is to better use existing classifiers and to design new models, based on the strong connection between GLR classifiers and DS theory laid bare in this paper. For instance, it would be interesting to study the properties of other decision rules in the belief function and imprecise probability frameworks, such as  maximality and e-admissibility \cite{troffaes07}. New classifier fusion schemes could be devised by combining the classifier output mass functions instead of  aggregating decisions by majority voting or averaging probabilities. And alternatives to Dempster's rule, such as the cautious rule \cite{denoeux08}, could be investigated,  to combine both feature-level mass functions inside the classifier, and output mass functions from a classifier ensemble.


%

\appendix

\section{Proof of Proposition \ref{prop:mass}}

\label{sec:mass}


\subsection{Expression of $m^+$}

As all positive masses $m_{k}^+$ defined  by (\ref{eq:mkp}) have the singletons $\stheta{k}$ and $\Theta$ as only focal elements, so has their orthogonal sum $m^+$. We thus have
\begin{multline*}
m^+(\stheta{k}) \propto \left[1-\exp\left(-w_{k}^+\right)\right] \prod_{l\neq k} \exp\left(-w_{l}^+\right)=\\
\prod_{l\neq k} \exp\left(-w_{l}^+\right) -\prod_{l=1}^K \exp\left(-w_{l}^+\right)=
 \left[\exp(w_k^+)-1\right] \exp\left(-\sum_{l=1}^K w_l^+\right)
\end{multline*}
and
$
m^+(\Theta) \propto  \exp\left(-\sum_{l=1}^K w_l^+\right).
$
Consequently,
\begin{equation*}
\sum_{k=1}^K m^+(\stheta{k}) + m^+(\Theta) \propto 
\exp\left(-\sum_{l=1}^K w_l^+\right)\left[\sum_{k=1}^K\exp(w_k^+) -K+1 \right]
\end{equation*}
and we have
\begin{eqs}{eq:mplus}
m^+(\stheta{k}) &= \frac{\exp(w_k^+)-1}{\sum_{l=1}^K\exp(w_l^+) -K+1} , \; k=1,\ldots,K\\
m^+(\Theta)&= \frac{1}{\sum_{l=1}^K\exp(w_l^+) -K+1}.
\end{eqs}
We note that $m^+(\stheta{k})$ is an increasing function of the total weight of evidence $w_k^+$ supporting $\theta_k$, and $m^+(\Theta)$ tends to one when all the positive weights $w_k^+$ tend to zero. 

\subsection{Expression of $m^-$}

The degree of conflict when combining the negative mass functions $m_k^-$, $k=1,\ldots,K$, defined by (\ref{eq:mkm}) is
\begin{equation}
\label{eq:kappam}
\kappa^- = \prod_{k=1}^K \left[1-\exp(-w_k^-)\right].
\end{equation}
We thus have, for any strict subset $A\subset \Theta$,
\begin{subequations}
\label{eq:mmoins}
\begin{equation}
m^-(A) =
 \frac{\left\{\prod_{\theta_k\not\in A} \left[1-\exp(-w_k^-)\right]\right\}\left\{\prod_{\theta_k\in A} \exp(-w_k^-)\right\}}{1-\prod_{k=1}^K \left[1-\exp(-w_k^-)\right]} , 
\end{equation}
and
\begin{equation}
m^-(\Theta)= \frac{ \exp\left(-\sum_{k=1}^K w_k^-\right)}{1-\prod_{k=1}^K \left[1-\exp(-w_k^-)\right]}.
\end{equation}
\end{subequations}
From (\ref{eq:plm} and  (\ref{eq:kappam}), the corresponding contour function is
\begin{equation}
\label{eq:plmoins}
pl^-(\theta_k)=\frac{\exp(-w_k^-)}{1-\prod_{l=1}^K \left[1-\exp(-w_l^-)\right]} , \quad k=1,\ldots,K.
\end{equation}

\subsection{Combination of $m^+$ and $m^-$}

Let
$
\eta^+ = \left(\sum_{l=1}^K\exp(w_l^+) -K+1\right)^{-1}
$
and
$
\eta^- = \left(1-\prod_{l=1}^K \left[1-\exp(-w_l^-)\right]\right)^{-1}.
$
From (\ref{eq:mplus}) and (\ref{eq:mmoins}), the degree of conflict between $m^-$ and $m^+$ is
\begin{align*}
\kappa &= \sum_{k=1}^K  \left\{m^+(\stheta{k}) \sum_{A \not\ni \theta_k} m^-(A)\right\}\\
&= \sum_{k=1}^K  \left\{m^+(\stheta{k}) (1-pl^-(\theta_k) \right\}\\
&= \sum_{k=1}^K  \left\{\eta^+\left(\exp(w_k^+)-1\right) [1-  \eta^-\exp(-w_k^-)] \right\}.
\end{align*}
Let $\eta=(1-\kappa)^{-1}$. We have, for any $k\in \{1,\ldots,K\}$,
\begin{multline*}
m(\stheta{k}) = 
\eta\left\{m^+(\stheta{k}) \left[\sum_{A \ni \theta_k} m^-(A)\right] + m^-(\stheta{k})m^+(\Theta)\right\}=\\
 \eta\left\{m^+(\stheta{k}) pl^-(\theta_k)+ m^-(\stheta{k})m^+(\Theta)\right\}.
\end{multline*}
Using Eqs.  (\ref{eq:mplus}), (\ref{eq:mmoins}) and (\ref{eq:plmoins}), we get
\begin{equation*}
m(\stheta{k})= \eta \eta^- \eta^+ \exp(-w_k^-)
\left\{ \exp(w_k^+)-1 + \prod_{l\neq k} \left[1-\exp(-w_l^-)\right]\right\}.
\end{equation*}
And for any $A\subseteq \Theta$ such that $|A|>1$,
\begin{equation*}
m(A) = \eta \; m^-(A) m^+(\Theta) =
\eta \; \eta^- \eta^+ \left\{\prod_{\theta_k\not\in A} \left[1-\exp(-w_k^-)\right]\right\}\left\{\prod_{\theta_k\in A} \exp(-w_k^-)\right\},
\end{equation*}
which completes the proof of Proposition \ref{prop:mass}.

\section{Proof of Proposition \ref{prop2}}
\label{sec:prop2}

 Developing the square in (\ref{eq:critere}), we get
\begin{equation}
\label{eq:fcalpha}
f(\bc,\balpha) =\sum_{j,k}  (\betah_{jk}+c_j)^2 \left(\sum_{i=1}^n \phi_j(x_{i})^2 \right)+ n\sum_{j,k} \alpha_{jk}^2  +
2 \sum_{j,k}(\betah_{jk}+c_j)\alpha_{jk}\sum_{i=1}^n \phi_j(x_{i}).  
\end{equation}
Assuming, as in Section \ref{subsec:ident2}, the features  $\phi_j$ to be centered, the last term in the right-hand side of (\ref{eq:fcalpha}) vanishes, and  we get
\begin{equation}
\label{eq:f}
f(\bc,\balpha) =\sum_{j=1}^J \left(\sum_{i=1}^n \phi_j(x_{i})^2 \right)\sum_{k=1}^K (\betah_{jk}+c_j)^2 + n\sum_{j,k} \alpha_{jk}^2. 
\end{equation}
Due to constraints (\ref{eq:constr}), for any $c_0$, the second term in the right-hand side of (\ref{eq:f}) is minimized for
\[
\alpha_{jk} = \frac{1}{J}(\betah_{0k} + c_0), \quad \text{for } j=1,\ldots,J \text{ and } k=1\ldots,K.
\]
Hence, the problem becomes
\begin{equation*}
\min_{\bc} f(\bc)=\sum_{j=1}^J \left(\sum_{i=1}^n \phi_j(x_{i})^2 \right) \left\{\sum_{k=1}^K (\betah_{jk}+c_j)^2 \right\}+ 
\frac{n}{J} \sum_{k=1}^K (\betah_{0k} + c_0)^2.
\end{equation*}
Each of the $J+1$ terms in this sum can be minimized separately. The solution can easily be found to be
\[
c^*_j=-\frac{1}{K} \sum_{k=1}^K \betah_{jk}, \quad  j=0,\ldots,J.
\]
The optimum coefficients are, thus,
\begin{equation}
\label{eq:beta}
\beta_{jk}^*=\betah_{jk} - \frac{1}{K} \sum_{l=1}^K\betah_{jl}, 
\end{equation}
for $j=0,\ldots,J$ and  $k=1\ldots,K$, and $\alpha_{jk}^* = \beta_{0k}^*/J$ for  $j=1,\ldots,J$ and $k=1\ldots,K$.

Let us now consider the case where the features are not centered. As before, let $\phi'_j(x_i)=\phi_j(x_i)-\mu_j$ denote  the centered feature values. We can write
\[
w_{ijk}=\beta_{jk} \phi_j(x_i)+ \alpha_{jk}=\beta_{jk} \phi'_j(x_i)+ \alpha'_{jk},
\]
with 
$
\alpha'_{jk}=\alpha_{jk} + \beta_{jk}\mu_j,
$
and
\[
\sum_{j=1}^J w_{ijk}= \sum_{j=1}^J\beta_{jk} \phi_j(x_i)+\beta_{0k}=\sum_{j=1}^J\beta_{jk} \phi'_j(x_i)+\beta'_{0k},
\]
with 
$
\beta'_{0k}=\beta_{0k} +\sum_{j=1}^J \beta_{jk}\mu_j.
$
The coefficients $\beta_{jk}$ are not modified, except for $j=0$. The optimal value of $\beta_{0k}$ is
\begin{align*}
\beta_{0k}'^*&=\betah'_{0k} - \frac{1}{K} \sum_{l=1}^K\betah'_{0l}\\
&=\betah_{0k} +\sum_{j=1}^J \betah_{jk}\mu_j - \frac{1}{K} \sum_{l=1}^K \left(\betah_{0l} +\sum_{j=1}^J \betah_{jl}\mu_j\right)\\
&=\betah_{0k}- \frac{1}{K} \sum_{l=1}^K \betah_{0l} +\sum_{j=1}^J \left(\betah_{jk}- \frac{1}{K} \sum_{l=1}^K \betah_{jl}\right) \mu_j \\
&=\betah_{0k}- \frac{1}{K} \sum_{l=1}^K \betah_{0l} +\sum_{j=1}^J \beta^*_{jk}\mu_j.
\end{align*}
Consequently, 
\[
\beta_{0k}^*=\beta_{0k}'^*-\sum_{j=1}^J \beta^*_{jk}\mu_j=\betah_{0k}- \frac{1}{K} \sum_{l=1}^K \betah_{0l}.
\] 
Now,
\[
\alpha_{jk}'^* = \beta_{0k}'^*/J=\frac1J\left(\beta_{0k}^*+\sum_{j=1}^J \beta^*_{jk}\mu_j\right).
\]
Hence,
\begin{equation*}
\alpha_{jk}^* =\alpha_{jk}'^* -\beta_{jk}^*\mu_j=\frac1J\left(\beta_{0k}^*+\sum_{j=1}^J \beta^*_{jk}\mu_j\right)-\beta_{jk}^*\mu_j,
\end{equation*}
which completes the proof.

\section*{Ackowledgements}

This research was supported by the Labex MS2T, which was funded by the French Government, through the program ``Investments for the future'' by the National Agency for Research (reference ANR-11-IDEX-0004-02).

\section*{References}


\end{document}